\documentclass{article}
\usepackage{ijcai23}

\usepackage{times}
\usepackage{soul}
\usepackage{url}
\usepackage{CJKutf8}

\usepackage[hidelinks]{hyperref}
\usepackage[utf8]{inputenc}
\usepackage[small]{caption}
\usepackage{graphicx}
\usepackage{amsmath}
\usepackage{amsthm}
\usepackage{booktabs}
\usepackage{algpseudocode}
\usepackage[switch]{lineno}
\usepackage{subcaption}

\usepackage[ruled,lined,linesnumbered]{algorithm2e}
\usepackage{multirow}

\usepackage{multicol,lipsum,xparse}
\usepackage{stmaryrd}
\usepackage{color}

\title{{Rubik's Optical Neural Networks}: \\
Multi-task Learning with Physics-aware Rotation Architecture}

\author{
Yingjie Li
\and
Weilu Gao\and
Cunxi Yu
\affiliations
$^1$University of Utah\\
\emails
\{yingjie.li, weilu.gao, cunxi.yu\}@utah.edu
}

\begin{document}

\maketitle

\begin{abstract}
Recently, there are increasing efforts on advancing optical neural networks (ONNs), which bring significant advantages for machine learning (ML) in terms of power efficiency, parallelism, and computational speed. With the considerable benefits in computation speed and energy efficiency, there are significant interests in leveraging ONNs into medical sensing, security screening, drug detection, and autonomous driving. However, due to the challenge of implementing reconfigurability, deploying multi-task learning (MTL) algorithms on ONNs requires re-building and duplicating the physical diffractive systems, which significantly degrades the energy and cost efficiency in practical application scenarios. This work presents a novel ONNs architecture, namely, \textit{RubikONNs}, which utilizes the physical properties of optical systems to encode multiple feed-forward functions by physically rotating the hardware similarly to rotating a \textit{Rubik's Cube}. To optimize MTL performance on RubikONNs, two domain-specific physics-aware training algorithms \textit{RotAgg} and \textit{RotSeq} are proposed. Our experimental results demonstrate more than 4$\times$ improvements in energy and cost efficiency with marginal accuracy degradation compared to the state-of-the-art approaches. 
\end{abstract}

\section{Introduction}

Recently, use of Deep Neural Networks (DNNs) shows significant advantages in many applications, including large-scale computer vision, natural language processing, and data mining tasks.  
However, DNNs have substantial computational and memory requirements, which greatly limit their training and deployment in resource-constrained (e.g., computation, I/O, and memory bounded) environments \cite{jouppi2017datacenter,yin2022exact,yazdanbakhsh2021evaluation,yin2023respect}. More importantly, it is identified that training large DNN models produces significant carbon dioxide, e.g., recent studies estimated 626,000 pounds of planet-warming carbon dioxide, equal to the lifetime emissions of five cars, produced in training Transformer network \cite{strubell2019energy}. As models grow bigger, their demand for computing increases, as well as the carbon footprint produced by those computations. To address these challenges and make the computation more eco-friendly, there has been a significant trend in building novel high-performance DNNs platforms, especially the increasing efforts on implementing novel DNNs in optical domain, i.e., optical neural network (ONNs) that mimic conventional feed-forward neural network functions using light propagation \cite{lin2018all,gu2020towards,ying2020electronic,shen2017deep,chen2022physics,chen2022complex,li2022physics,li2021late,duan2023optical}. Unlike directly accelerating conventional DNNs, algorithms for training and deploying ONNs need to be customized in order to precisely represent the whole physics properties of light propagation. Specifically, the equivalent numerical representations of inputs, intermediate results, and propagation functions in optical domain are complex values and complex-valued functions. 
{Additionally, due to the limitations from nature physics, implementing reconfigurability and deploying multi-task learning (MTL) algorithms on many ONNs systems requires re-building and duplicating the physical hardware systems, which significantly degrades the energy and cost efficiency in practical application scenarios. 
}

This work proposes a novel architecture \textbf{RubikONNs}, which utilizes the physical properties of optical systems to encode multiple feed-forward functions by physically rotating the systems similarly as rotating a \textit{Rubik's Cube}. {With the realization of MTL in optical systems, the computational carbon footprint can be significantly reduced while maintaining the system performance.} {The paper is organized as follows: in Section \ref{sec:background}, we introduce Diffractive Deep Neural Networks (D$^2$NN) and its physical implementations; in Section \ref{sec:approch}, we first formulate the forward functions in D$^2$NN systems for MTL. Furthermore, to optimize the MTL performance of RubikONNs, we propose two novel domain-specific physics-aware training algorithms, \textbf{RotAgg} and \textbf{RotSeq}; in Section \ref{sec:results}, we demonstrate four-task MTL on RubikONNs with implementation cost and energy efficiencies improved more than \textbf{4$\mathbf{\times}$}. Finally, a comprehensive RubikONNs design space exploration analysis and explainability are provided to offer concrete design methodologies for practical uses.}

\begin{figure*}[t]
    \centering
    \includegraphics[width=0.93\linewidth]{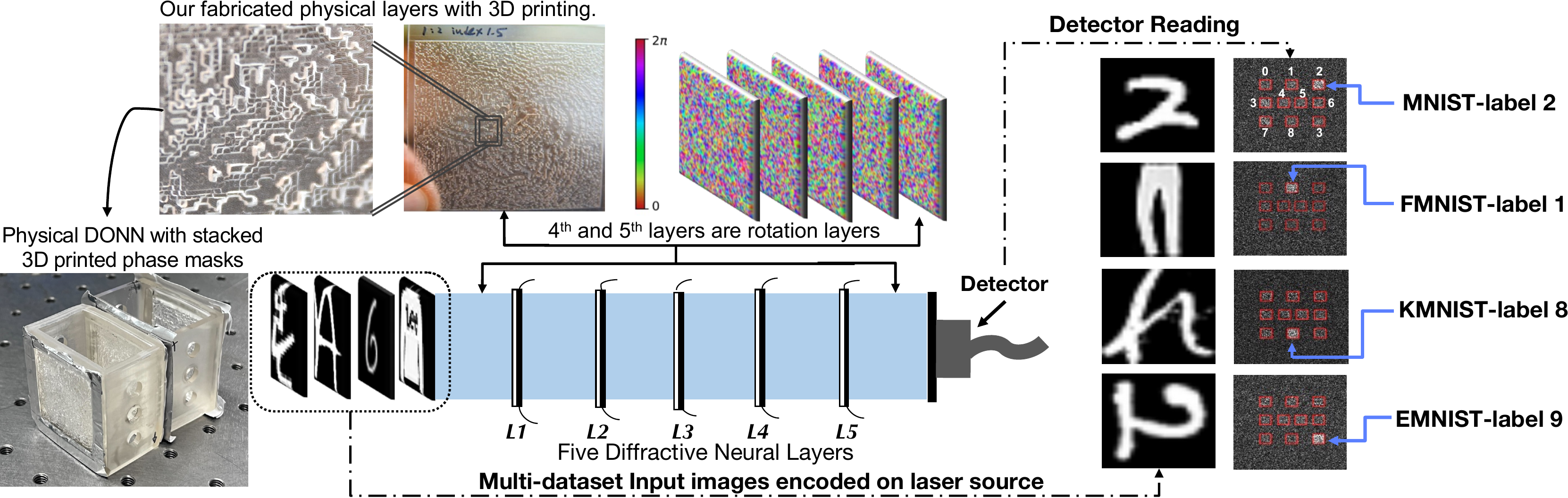}
    \caption{Overview of Rubik DONN system, consisting of (1) laser source that encodes the input images, (2) diffractive layers with trainable phase parameters (weights), which are the non-configurable passive optical devices fabricated with 3D printing and stacked together as DONN systems, and (3) detectors that reads the output. The four inference results are collected with specific rotations patterns shown in Figure \ref{fig:rotation}.}
   \vspace{-3mm}
    \label{fig:system}
\end{figure*}

\section{Background}
\label{sec:background}
\noindent\textbf{Diffractive Deep Neural Networks (D$^2$NN)} -- {Recently, there are increasing efforts on optical neural networks and optical computing based DNNs hardware, which bring significant advantages for machine learning systems in terms of their power efficiency, parallelism, and computational speed, demonstrated at various optical computing systems by \cite{mengu2020scale,lin2018all,feldmann2019all,shen2017deep,tait2017neuromorphic,rahman2020ensemble,li2020multi,gu2022adept,gao2021artificial,tang2023device,lou2023effects}. Among them, free-space \textit{diffractive deep neural networks} (D$^2$NNs) , which is based on the light diffraction and phase modulation of the light signal provided by diffractive layers (L1-L5 in Figure \ref{fig:system}), featuring millions of neurons in each layer interconnected with neurons in neighboring layers. This ultrahigh density and parallelism make this system possess fast and high throughput computing capability. {{Additionally, the D$^2$NN system is implemented with passive optical devices, where the devices function without additional maintaining power required, thus significantly reducing the consumption power for solving deep learning tasks with orders of magnitude energy efficiency advantages over low-power digital devices (\cite{lin2018all,rahman2020ensemble,mengu2023snapshot,li2020multi,chen2022physics,li2022physics}).}} More importantly, \cite{lin2018all,rahman2020ensemble,li2020multi,chen2022physics,mengu2020scale,li2022physics} demonstrated that diffractive propagation controlled by phase modulation are differentiable, which means that such parameters can be optimized with conventional backpropagation algorithms using conventional automatic differentiation (\texttt{autograd}) engine implemented in {modern compilers such as }PyTorch and Tensorflow. 

In conventional DNNs, forward propagation is computed by generating the feature representation with floating-point weights associated with each neural layer. While in D$^2$NNs, such floating-point weights are encoded in the phase modulation of each neuron in diffractive phase masks, which is acquired by and multiplied onto the light wavefunction as it propagates through the neuron. Similar to conventional DNNs, the final output class is predicted based on generating labels according to a given one-hot representation, e.g., the max energy reading over the output signals of the last layer observed by detectors. Specific examples of the system at training and inference can be found in the next section. 

Once the training of a D$^2$NN system is completed on the digital computation platform, the trained D$^2$NN is deployed on the optical platform with non-configurable fabricated phase masks such as 3D printed phase masks, as diffractive layers for all-optical inference. Thus, D$^2$NNs lack reconfigurability for the weight parameters, which will bring significant energy and system cost overhead in practical application scenarios, especially for MTL.
}

\section{Approach}\label{sec:approch}

To overcome the aforementioned limitations of existing D$^2$NN systems, we propose a novel neural architecture, namely \textbf{RubikONNs}, that utilizes the physical rotation properties of existing D$^2$NN systems to realize MTL with few overheads, in which case, a single-task D$^2$NN system can be used to encode multiple feed-forward functions by rotating its underlying structure just like rotating a \textit{Rubik's Cube}.

\begin{figure}[t]
    \centering
    \includegraphics[width=1\linewidth]{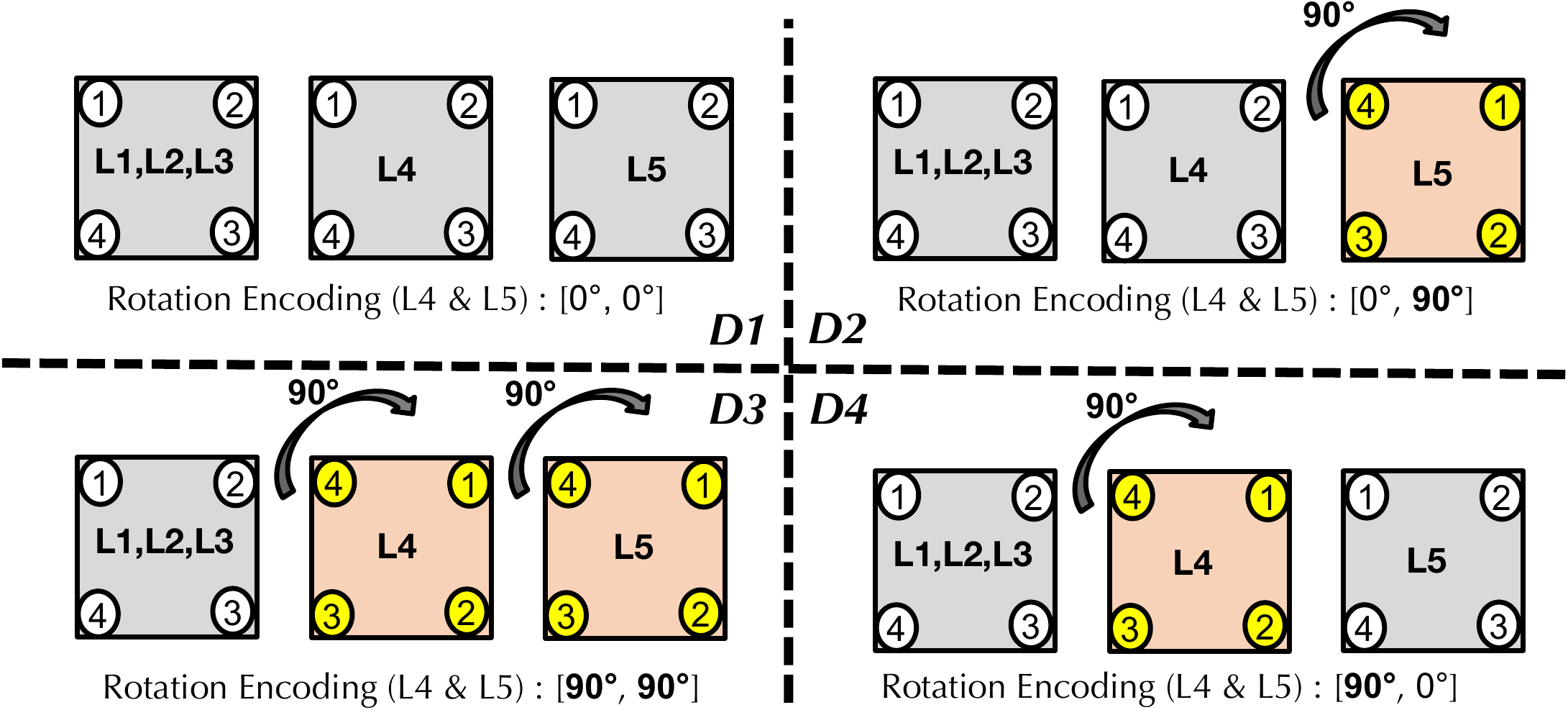}
    \caption{Example of a four-task RubikONNs rotation architecture.}
    \label{fig:rotation}
    \vspace{-3mm}
\end{figure}
 
\subsection{Forward function for a single-task D$^2$NN}

D$^2$NN system is designed with three major components (Figure \ref{fig:system}): (1) laser source encoding the input images, (2) diffractive layers encoding trainable phase modulation, and (3) detectors capturing the output of the forward propagation. Specifically, the input image is first encoded with the laser source. The information-encoded light signal is diffracted in the free space between diffractive layers and modulated via phase modulation at each layer. Finally, the diffraction pattern after light propagation w.r.t light intensity distribution will be captured at the detector plane for predictions. 

From the beginning of the system, the input information (e.g., an image) is encoded on the coherent light signal from the laser source, its wavefunction can be expressed as $f^{0}(x_{0}, y_{0})$. The wavefunction after light diffraction from the input plane to the first diffractive layer over diffraction distance $z$ can be seen as the summation of the outputs at the input plane, i.e., 
\begin{equation}
\label{eq:diffraction_time}
    f^{1}(x, y) = \iint f^{0}(x_{0}, y_{0})h(x-x_{0}, y-y_{0}, z)dx_{0}dy_{0}
\end{equation}
where $(x, y)$ is the coordinate on the receiver plane, i.e., the first diffractive layer, $h$ is the impulse response function of free space. Here we use Fresnel approximation, thus the impulse response function $h$ is
\begin{equation}
\label{eq:diffraction_time_end}
    h(x, y, z) = \frac{\exp(ikz)}{i\lambda z}\exp\{\frac{ik}{2z}(x^{2} + y^{2})\}
\end{equation}
where $i=\sqrt{-1}$, $\lambda$ is the wavelength of the laser source, $k=2\pi/\lambda$ is free-space wavenumber.

Equation \ref{eq:diffraction_time} can be calculated with spectral algorithm, where we employ Fast Fourier Transform (FFT) for fast and differentiable computation, i.e., 
\begin{equation}
\label{equ:diffraction_freq}
    U^{1}(\alpha, \beta) = U^{0}(\alpha, \beta)H(\alpha, \beta, z)
\end{equation}
where $U$ and $H$ are the Fourier transformation of $f$ and $h$ respectively. 

After light diffraction, the wavefunction resulting in Equation \ref{equ:diffraction_freq} $U^{1}(\alpha, \beta)$ is first transformed to time domain with inverse FFT (iFFT). Then the phase modulation $W(x, y)$ provided by the diffractive layer is applied to the light wavefunction in time domain by matrix multiplication, i.e.,
\begin{equation}
\label{eq:phase_mod}
    f^{2}(x, y) = \text{iFFT}(U^{1}(\alpha, \beta)) \times W_{1}(x, y)
\end{equation}
where $W_{1}(x, y)$ is the phase modulation in the first diffractive layer, $f^{2}(x, y)$ is then the input light wavefunction for the light diffraction between the first diffractive layer and the second diffractive layer. 

We enclose one computation round of light diffraction and phase modulation at one diffractive layer as a computation module named \textbf{DiffMod}, i.e.,
\begin{equation}
\label{eq:diffmod}
    \text{DiffMod}(f(x, y), W) = L(f(x, y), z) \times W(x, y)
\end{equation}
where $f(x, y)$ is the input wavefunction, $W(x, y)$ is the phase modulation, $L(f(x, y), z)$ is the wavefunction after light diffraction over a constant distance $z$ in time domain, i.e., $\text{iFFT}(U(\alpha, \beta))$ in Equation \ref{eq:phase_mod}.  

As a result, in a multiple diffractive layer constructed D$^2$NN system, the forward function can be computed iteratively for the stacked diffractive layers. For example, for the 5-layer system shown in Figure \ref{fig:system}, the forward function can be expressed as,
\begin{equation}\label{eq:forward}
\small
    \begin{split}
       I(f^{0}(x, y), W) = \text{DiffMod}(\text{DiffMod}(\text{DiffMod}(\text{DiffMod}\\(
       \text{DiffMod}(f^{0}(x, y),
       W_{1}(x, y)),W_{2}(x, y)), \\
       W_{3}(x, y)), W_{4}(x, y)), W_{5}(x, y))
    \end{split}
\end{equation}
where $f^{0}(x, y)$ is the input wavefunction to the system and $W_{1-5}$ is phase modulation provided at each diffractive layer. 

The final diffraction pattern w.r.t the light intensity $I$ in Equation \ref{eq:forward} is projected to the detector plane. {We can design arbitrary detector patterns for classes in different tasks by setting the corresponding coordinates of the detector region at the full detector plane for each class by the user's definition. For example,} for MNIST datasets, the output plane is divided into \textbf{ten} detector regions to mimic the output of conventional neural networks for predicting \textbf{ten} classes. The final class will be produced by \texttt{argmax} function with the ten intensity sums of the ten detector regions as input. For example, in Figure \ref{fig:system}, based on the label indices of the ten detector regions for image "2", we can see that the 3$^{rd}$ region on the first row has the highest energy. Then, the predicted class is class "2". Similarly, the predicted classes "1", "8", and "9" of other three datasets can be generated by applying \texttt{argmax} on the detector. With the one-hot represented ground truth class $t$, the loss function $L$ can be acquired with \textbf{MSELoss} as,
\begin{equation}
    L = \parallel \text{Softmax}(I) - t \parallel_{2}
\end{equation}
Thus, the whole system is designed to be differentiable and compatible with conventional automatic differential engines.

\subsection{RubikONNs Architecture for MTL}

To deal with multiple tasks with minimum system overhead, an ideal system should be designed to encode different forward functions without changing the single-task system. Note that the diffractive layers are mostly designed with 3D printed materials, such that the phase parameters (weights) carried by these layers are non-reconfigurable after 3D printing. However, as demonstrated by \cite{lin2018all,li2020multi}, the layers are portable in D$^2$NNs and they are in square shapes. This means that we are able to rotate each layer by close-wise $90^\circ$, $180^\circ$, or $270^\circ$, and place the layer back in the system without any other changes. While each layer carries specific trained phase parameters, by rotating one or multiple layers, the forward function will be different since the weights of the model are changed. In optical domain, this means that the modulation of the light changes accordingly w.r.t specific rotation patterns. This offers the main motivation of designing RubikONNs that aims to enable MTL in existing single-task D$^2$NN systems. As a result, RubikONNs enables MTL by simply \textbf{(1)} pulling out the layer, \textbf{(2)} rotating it to the specific rotation pattern as designed, and \textbf{(3)} plugging the layer back to the original location, without changing the rest of the system.

To illustrate the rotation architecture RubikONNs, an example of encoding four tasks with the last two layers (L4, L5) as \textit{rotation layers} is shown in Figure \ref{fig:rotation}. The designed rotation type applied to these layers is the \textit{rotation angle} (clockwise $90^\circ$ in this example). 
{
To summarize the forward function of RubikONN architecture, we introduce two Boolean variables to indicate the rotation patterns of L4 and L5 layers. When $s_{0}=1$, L4 will be rotated clockwise $90^\circ$, otherwise, L4 will remain unchanged; similarly, $s_{1}$ indicates the rotation pattern of L5. Thus, for the first task, $s_{0}s_{1}=00$, both layers are unchanged; for the second task, $s_{0}s_{1}=01$, L5 will be rotated clockwise $90^\circ$; for the third task, $s_{0}s_{1}=11$, both layers will be rotated clockwise $90^\circ$; for the fourth task, $s_{0}s_{1}=10$, L4 will be rotated clockwise $90^\circ$. The forward function is expressed as follows:
}
\begin{equation}
\scriptsize
I = \text{DiffMod}(\text{DiffMod}(\underset{\substack{1\leq i \leq3}}I(f^{0}, W_{i}), 
    \begin{cases}
    W_{4}), W_{5}),&{s_{0}s_{1} = 00}\\
    W_{4}), \text{Rot}(W_{5})),&{s_{0}s_{1} = 01} \\
    \text{Rot}(W_{4})), \text{Rot}(W_{5})),&{s_{0}s_{1} = 11}\\
    \text{Rot}(W_{4})), W_{5}),&{s_{0}s_{1} = 10} \\
    \end{cases}
\end{equation}

{We take $s_{0}s_{1} = 01$ as an example for the quantitative analysis w.r.t the rotation in the system, where only the fifth layer is rotated for 90$^{\circ}$. Thus, the forward function is 
\begin{equation}
    I_{01} = \text{DiffMod}(((\underset{\substack{1\leq i \leq4}}I(f^{0}, W_{i}), \text{Rot}(W_{5}))
\end{equation}
where $\text{DiffMod}(f(x, y), W) = L(f(x, y), z) \times W(x, y), x \in [n, n], y \in [n, n]$ as shown in Equation \ref{eq:diffmod}, assuming $W_{5}$ is trained as $W_{n, n}$ and the diffraction result is $L_{n, n}$, i.e., 
\begin{align*}
\footnotesize
W_{n,n} = 
\begin{pmatrix}
w^{1}_{1} & w^{1}_{2} & \cdots & w^{1}_{n} \\
w^{2}_{1} & w^{2}_{2} & \cdots & w^{2}_{n} \\
\vdots  & \vdots  & \ddots & \vdots \\
w^{n}_{1} & w^{n}_{2} & \cdots & w^{n}_{n} 
\end{pmatrix}
&
L_{n,n} = 
\begin{pmatrix}
l^{1}_{1} & l^{1}_{2} & \cdots & l^{1}_{n} \\
l^{2}_{1} & l^{2}_{2} & \cdots & l^{2}_{n} \\
\vdots  & \vdots  & \ddots & \vdots  \\
l^{n}_{1} & l^{n}_{2} & \cdots & l^{n}_{n} 
\end{pmatrix}
\end{align*}

Thus, the wavefunction after DiffMod($f, W$) \textbf{without rotation}, i.e., $I_{00}$, is calculated as
\begin{equation*}
\footnotesize
I_{00} = 
\begin{pmatrix}
l^{1}_{1} * w^{1}_{1} & l^{1}_{2} * w^{1}_{2} & \cdots & l^{1}_{n} * w^{1}_{n} \\
l^{2}_{1} * w^{2}_{1} & l^{2}_{2} * w^{2}_{2} & \cdots & l^{2}_{n} * w^{2}_{n} \\
\vdots  & \vdots  & \ddots & \vdots  \\
l^{n}_{1} * w^{n}_{1} & l^{n}_{2} * w^{n}_{2} & \cdots & l^{n}_{n} * w^{n}_{n}
\end{pmatrix}
\end{equation*}

When the rotation pattern applied to $W_{5}$ is $90 ^\circ$, the corresponding $\text{Rot}(W_{5})$is

\begin{equation*}
\footnotesize
\text{Rot}(W_{5}) = 
\begin{pmatrix}
w^{n}_{1} & \cdots & w^{2}_{1} & w^{1}_{1}\\
w^{n}_{2} & \cdots & w^{2}_{2} & w^{1}_{2}\\
\vdots  & \ddots & \vdots   & \vdots  \\
w^{n}_{n} & \cdots & w^{2}_{n} & w^{1}_{n} 
\end{pmatrix}
\end{equation*}
While $L(f, z)$ remains the same, the correponding rotated $I_{01}$ is thus altered to 

\begin{equation*}
\footnotesize
I_{01} = 
\begin{pmatrix}
l^{1}_{1} * w^{n}_{1} & \cdots & l^{1}_{n-1} * w^{2}_{1} & l^{1}_{n} * w^{1}_{1} \\
l^{2}_{1} * w^{n}_{2} & \cdots & l^{2}_{n-1} * w^{2}_{2} & l^{2}_{n} * w^{1}_{2} \\
\vdots & \ddots & \vdots   & \vdots  \\
l^{n}_{1} * w^{n}_{n} & \cdots & l^{n}_{n-1} * w^{2}_{n} & l^{n}_{n} * w^{1}_{n}
\end{pmatrix}
\end{equation*}

As a result, by rotating the weights matrix $W$ (phase modulation) to different angles, the information-carried light signal is modulated with different applied phase modulations w.r.t different datasets. The full propagation figures in Figure \ref{fig:prop-mnist} show the light patterns when different rotation patterns are applied to the same input light signal.
}

Note that the proposed architecture can go beyond four tasks by adding more rotation patterns. For example, each layer has four rotation states, $0^\circ$, $90^\circ$, $180^\circ$, or $270^\circ$, such that the maximum number of different forward functions is $1 + 3\times4 = 13$ {with two rotation layers; }{{when add another rotation layer, i.e., three rotation layers with four rotation states in the system, the system can deal with up to 29 tasks.}}. Discussions about choices of rotation layers and different rotation angles are included in Section \ref{sec:results}. 

While the RubikONNs architecture enables zero-overhead MTL on D$^2$NN systems, the training algorithms that are aware of physical rotations and light diffraction do not exist. Specifically, the training algorithms should be able to learn structural weight parameters w.r.t specific rotation patterns and given datasets. Thus, we introduce two novel MTL algorithms for training RubikONNs, i.e., \textit{rotated aggregation algorithm (RotAgg)} and \textit{rotated sequence Algorithm (RotSeq)}.

\begin{algorithm}[t]
\scriptsize
\SetAlgoLined
\KwResult{$\boldsymbol{W} = \{\boldsymbol{W}_{C}^{1,2,3},\boldsymbol{W}_{R}^{4},~\boldsymbol{W}_{R}^{5}\}$ for the rotation model}

\textbf{Initialization}: Weights $\boldsymbol{W_{0}}^{0} =\{ \boldsymbol{W_{C_{0}}}^{1,2,3} , \boldsymbol{W_{R_{0}}}^{4} , \boldsymbol{W_{R_{0}}}^{5}\}$ for the model \algorithmiccomment{Weights initialization} \label{lst:avg:init} 

 \While{i $\leq$ training iterations}{
  $\boldsymbol{W_{1,2,3,4}}^{i} \leftarrow \boldsymbol{W_{0}}^{i}$\label{lst:avg:reload} ; \\
   $\boldsymbol{W_{1}}^{i} \leftarrow \{ \boldsymbol{W}_{C_{i}}^{1, 2, 3} , \boldsymbol{W}_{R_{i}}^{4}  \boldsymbol{W}_{R_{i}}^{5}\}$;
   $\boldsymbol{W_{1}}^{i+1} \xleftarrow{D1} \boldsymbol{W_{1}}^{i} - \eta\nabla \boldsymbol{W_{1}}^{i}$;\label{lst:avg:w1update}  \algorithmiccomment{training w.r.t task 1 (D1) w/o rotation.} \\
   
   $\boldsymbol{W_{2}}^{i} \leftarrow\{ \boldsymbol{W}_{C_{i}}^{1, 2, 3}, \boldsymbol{W}_{R_{i}}^{4} ,\texttt{rotate} (\boldsymbol{W}_{R_{i}}^{5})\}$;
   $\boldsymbol{W_{2}}^{i+1} \xleftarrow{D2} \boldsymbol{W_{2}}^{i} - \eta\nabla \boldsymbol{W_{2}}^{i}$; \label{lst:avg:w2update}  \algorithmiccomment{task 2 (D2) update w 5$^{th}$ layer rotated $90^\circ$}\\
   
    $\boldsymbol{W_{3}}^{i} \leftarrow \{ \boldsymbol{W}_{C_{i}}^{1, 2, 3} , \texttt{rotate} (\boldsymbol{W}_{R_{i}}^{4}) , \texttt{rotate}(\boldsymbol{W}_{R_{i}}^{5})\}$;
   $\boldsymbol{W_{3}}^{i+1} \xleftarrow{D3} \boldsymbol{W_{3}}^{i} - \eta\nabla \boldsymbol{W_{3}}^{i}$;\label{lst:avg:w3update} \algorithmiccomment{task 3 (D3) update w 4,5$^{th}$ layer rotated $90^\circ$.} \\
   
   $\boldsymbol{W_{4}}^{i} \leftarrow \{ \boldsymbol{W}_{C_{i}}^{1, 2, 3} , \texttt{rotate} (\boldsymbol{W}_{R_{i}}^{4}) , \boldsymbol{W}_{R_{i}}^{5}\}$;
   $\boldsymbol{W_{4}}^{i+1} \xleftarrow{D4} \boldsymbol{W_{4}}^{i} - \eta\nabla \boldsymbol{W_{4}}^{i}$; \label{lst:avg:w4update}  \algorithmiccomment{task 4 (D4) update w 4$^{th}$ layer rotated $90^\circ$.}  \\
    
$\boldsymbol{W_{2}}^{i} \leftarrow \{ \boldsymbol{W}_{C_{i}}^{1, 2, 3} , \boldsymbol{W}_{R_{i}}^{4} , \texttt{rotate-back} (\boldsymbol{W}_{R_{i}}^{5})\}$;
$\boldsymbol{W_{3}}^{i} \leftarrow \{ \boldsymbol{W}_{C_{i}}^{1, 2, 3} , \texttt{rotate-back} (\boldsymbol{W}_{R_{i}}^{4}) , \texttt{rotate-back}(\boldsymbol{W}_{R_{i}}^{5})\}$;
$\boldsymbol{W_{4}}^{i} \leftarrow \{ \boldsymbol{W}_{C_{i}}^{1, 2, 3} , \texttt{rotate} (\boldsymbol{W}_{R_{i}}^{4}) , \boldsymbol{W}_{R_{i}}^{5}\}$; \label{lst:avg:rotateback} \algorithmiccomment{reversely rotating virtual models for aggregation.} \\

   $\boldsymbol{W_{0}}^{i+1} \leftarrow$ $(\boldsymbol{W_{1}^{i+1}} + \boldsymbol{W_{2}^{i+1}} + \boldsymbol{W_{3}^{i+1}} + \boldsymbol{W_{4}^{i+1}})/4$; \label{lst:avg:avg}   \\
   $i \leftarrow i+1$
 }  
\caption{Rotated Aggregation Algorithm (\textbf{RotAgg}) for training RubikONNs.}
   \label{Alg: Averaging-weights}
\end{algorithm}

\subsection{Algorithm 1: Rotated Aggregation Training}

The Rotated Aggregation Training (\textbf{RotAgg}) algorithm shown in Alg. \ref{Alg: Averaging-weights} aims to update the parameters of RubikONNs by averagely aggregating gradients generated from all tasks, while the gradient of each task is computed by including the rotations in every training iteration. Therefore, the training iterations are fully aware of physical rotations of the RubikONNs architecture. We illustrate RotAgg using the same rotation designs shown in Figure \ref{fig:rotation}, where the first three layers are shared layers, named as $W_C^{1,2,3}$, that will not be rotated during training and inference, and the rotations layers are denoted as $W_R^{4}$ and $W_R^{5}$. First, RotAgg algorithm initializes one model for aggregation, and four virtual models to store temporary updates w.r.t specific rotation patterns and dataset (line~\ref{lst:avg:init}). In every training iteration, RotAgg will first update the parameters in the four virtual models, $W_{1,2,3,4}^{i}$, where the four models are trained separately w.r.t the designed rotation patterns and the corresponding dataset (lines~\ref{lst:avg:w1update} - \ref{lst:avg:w4update}). Note that at each iteration, the virtual model will be re-initialized before any gradient update, with the initial weight parameters or parameters optimized in the previous iteration (line~\ref{lst:avg:reload}). For example, the first update is performed for task 1 w.r.t dataset D1 (line~\ref{lst:avg:w1update}), where the model is rotated based on $W_1^{i}$ with the rotation pattern of $[0^\circ, 0^\circ]$ as shown in Figure \ref{fig:rotation}. The second update will then be performed w.r.t to task 2 dataset D2, where the virtual model $W_2^i$ will be initialized by rotating parameters in rotation layers (L4 and L5) with the rotation pattern $[0^\circ, 90^\circ]$ (line~\ref{lst:avg:w2update}). Similarly, the virtual models for task 3 (D3) and task 4 (D4) will be performed. Before the final weight aggregation, the four virtual models will be reverse-rotated back to the initial position (line~\ref{lst:avg:rotateback}). Finally, RotAgg averagely aggregates the weights from all four virtual models (line~\ref{lst:avg:avg}), and return $W_0^{i+1}$ for next iteration or as final model.

\subsection{Algorithm 2: Sequential Rotation Training }

The second training algorithm Sequential Rotation Training (\textbf{RotSeq}) shown in Alg. \ref{Alg: Sequencing-training} aims to update the parameters of RubikONNs by sequentially updating the model w.r.t a given sequence of task orders in order to incorporate the physical rotations in the training process. Here, we illustrate Alg. \ref{Alg: Sequencing-training} using a specific order of updates, i.e., D1$\rightarrow$D2$\rightarrow$D3$\rightarrow$D4.
In the illustration example, for the first task, the model will be updated w.r.t dataset D1 without rotating the rotation layers (line~\ref{lst:seq:w1update}). Unlike the RotAgg algorithm, the model will be directly updated to $W^i$ after the training of the first task. Next, the weights will be rotated with the rotation pattern $[0^\circ, 90^\circ]$, i.e., rotating $W_{R_i}^5$ clockwise $90^\circ$ before the gradient update process for task 2 (line~\ref{lst:seq:rotatew2}). Note that the model rotated before training for task 2 has already been updated w.r.t D1. Similarly, the model will be trained in the same sequential updating fashion according to the rotation patterns designed for task 3 (line~\ref{lst:seq:rotatew3}) and task 4 (line~\ref{lst:seq:rotatew4}). {\color{black}Note that the inner loop update order can be fixed for all iterations or can be dynamically changed through the training process.} Therefore, in addition to other training parameters, RotSeq could also be impacted by the inner loop update orders. In Section \ref{sec:results}, a comprehensive analysis of the update orders is provided.

\begin{algorithm}[t]
\scriptsize
\SetAlgoLined
\KwResult{$\boldsymbol{W} = \{\boldsymbol{W}_{C}^{1,2,3},\boldsymbol{W}_{R}^{4}, \boldsymbol{W}_{R}^{5}\}$ for the rotation model}
\textbf{initialization}: Weights $\boldsymbol{W}^{0}$ for the model; \label{lst:seq:init}\\
 \While{i $\leq$ training iterations}{
   $\boldsymbol{W}^{i} = \{\boldsymbol{W}_{C_{i}}^{1, 2, 3}, \boldsymbol{W}_{R_{i}}^{4} , \boldsymbol{W}_{R_{i}}^{5}\}$;\\

   $\boldsymbol{W}^{i} \xleftarrow{D1} \boldsymbol{W}^{i} - \eta\nabla \boldsymbol{W}^{i}$;\label{lst:seq:w1update} \\
   $\boldsymbol{W}^{i} \leftarrow \{\boldsymbol{W}_{C_{i}}^{1, 2, 3}, \boldsymbol{W}_{R_{i}}^{4} , \texttt{rotate}(\boldsymbol{W}_{R_{i}}^{5})\}$;\label{lst:seq:rotatew2} \algorithmiccomment{re-training w.r.t task 2 (D2) w 5$^{th}$ layer rotated $90^\circ$.}\\

   $\boldsymbol{W}^{i} \xleftarrow{D2} \boldsymbol{W}^{i} - \eta\nabla \boldsymbol{W}^{i}$; \label{lst:seq:w2update} \\
   $\boldsymbol{W}^{i} \leftarrow \{\boldsymbol{W}_{C_{i}}^{1, 2, 3}, \texttt{rotate}(\boldsymbol{W}_{R_{i}}^{4}) , \boldsymbol{W}_{R_{i}}^{5}\}$;\label{lst:seq:rotatew3} \algorithmiccomment{re-training w.r.t task 3 (D3) w 4$^{th}$ \& 5$^{th}$ layers rotated $90^\circ$.}\\

   $\boldsymbol{W}^{i} \xleftarrow{D3} \boldsymbol{W}^{i} - \eta\nabla \boldsymbol{W}^{i}$; \label{lst:seq:w3update}\\
   $\boldsymbol{W}^{i} \leftarrow \{\boldsymbol{W}_{C_{i}}^{1, 2, 3}, \boldsymbol{W}_{R_{i}}^{4},\texttt{rotate-back}(\boldsymbol{W}_{R_{i}}^{5})\}$;\label{lst:seq:rotatew4} \algorithmiccomment{re-training w.r.t task 4 (D4) w 4$^{th}$ layer rotated $90^\circ$.}\\

   $\boldsymbol{W}^{i} \xleftarrow{D4} \boldsymbol{W}^{i} - \eta\nabla \boldsymbol{W}^{i}$; \label{lst:seq:w4update}\\
   
   $\boldsymbol{W}^{i} \leftarrow \{\boldsymbol{W}_{C_{i}}^{1, 2, 3}, \texttt{rotate-back}(\boldsymbol{W}_{R_{i}}^{4}),\boldsymbol{W}_{R_{i}}^{5}\}$; \label{lst:seq:rotate-back} \algorithmiccomment{rotate back the original pattern for task 1 (D1)}\\

   $\boldsymbol{W}^{i+1} \leftarrow \boldsymbol{W}^{i}$; \label{lst:seq:wupdate}  \\
   $i \leftarrow i+1$
 }  
 \caption{Sequential Rotation Training Algorithm (\textbf{RotSeq}) for training RubikONNs.}
 \label{Alg: Sequencing-training}

\end{algorithm}
\vspace{-3mm}
 \section{Results}\label{sec:results}
\begin{table*}[h]
  \captionsetup{justification=centering}
  \caption{Evaluations of prediction performance on four-task multi-task learning using datasets. MNIST(D1), FMNIST(D2), KMNIST(D3), and EMNIST(D4), optimized with the proposed RotSeq and RotAgg algorithms.}
   \vspace{2mm}
  \label{table:permutation_result}
  \centering
    \resizebox{1.7\columnwidth}{!}{%
  \begin{tabular}{cccccc}
    \toprule

    Algorithm & Permutation & MNIST{\tiny D1}   & FMNIST{\tiny D2}    &KMNIST{\tiny D3}  &EMNIST{\tiny D4} \\
    \midrule
    \multirow{4}{*}{RotSeq (Alg. \ref{Alg: Sequencing-training})} & {D1 $\shortrightarrow$ D2 $\shortrightarrow$ D3 $\shortrightarrow$ \textbf{D4}}      &{0.9533} & {0.8384} & {0.8275} & {\bf 0.8885}   \\
    \multirow{4}{*} & {D4 $\shortrightarrow$ D1 $\shortrightarrow$ D2 $\shortrightarrow$ \textbf{D3}}     &{0.9539} & {0.8394} & {\bf 0.8353} & {0.8811}   \\
    \multirow{4}{*} & {D3 $\shortrightarrow$ D4 $\shortrightarrow$ D1 $\shortrightarrow$ \textbf{D2}}     &{0.9539} & {\bf 0.8430} & {0.8270} & {0.8824}   \\
    \multirow{4}{*} & {D2 $\shortrightarrow$ D3 $\shortrightarrow$ D4 $\shortrightarrow$ \textbf{D1}}    &{\bf 0.9558} & {0.8386} & {0.8272} & {0.8834}   \\
    %
    \hline 
    
    {RotAgg (Alg. \ref{Alg: Averaging-weights})} & n/a &{0.9524} &{0.8412} &{0.8272} &{0.8819}   \\ 
  \hline\hline
  
    \cite{li2020multi} & n/a  & {0.9532} & {0.8370} & {0.8277} & {0.8464}   \\  
    
    {BaselineMTL}  & n/a  & {0.9262} & {0.8176} & {0.7430} & {0.8174}   \\ 
    \bottomrule
  \end{tabular}
     }
\end{table*}

{
\noindent\textbf{System Setup} -- The model explored in our experiments is designed with five diffractive layers as it is shown in Figure \ref{fig:system}. The system size is set to be $200\times200$, i.e., the size of the diffractive layers and the total detector plane is $200\times200$. The input image whose original size is $28\times28$ will be enlarged to $200\times200$ and encoded on the laser light signal with the \textit{wavelength} = 532 $nm$. The physical distances between layers, first layer to source, and final layer to detector, are set to be $30~cm$. {The architecture is designed with the rotation patterns shown in Figure \ref{fig:rotation}.} 
The detector collects the light intensity of the ten pre-defined regions for ten classes with each size of $20\times20$ (Figure \ref{fig:system}), where the sums of intensity of these ten regions are equivalent to a 1$\times$10 vector in \texttt{float32} type. The final prediction results will be then generated using \texttt{argmax}. 

\noindent
\textbf{Training Parameters and Datasets} -- To evaluate the proposed RubikONNs architecture and RotAgg and RotSeq training algorithms, we select four public image classification datasets, including 1) \textit{MNIST-10} (MNIST) (\cite{lecun1998mnist},) 2) \textit{Fashion-MNIST} (FMNIST) (\cite{xiao2017fashion}), 3) \textit{Kuzushiji-MNIST} (KMNIST) (\cite{clanuwat2018deep}), and 4) \textit{Extension-MNIST-Letters} (EMNIST) (\cite{cohen2017emnist}), an extension of MNIST to handwritten letters. Specifically, for EMNIST, we customize the dataset to have the first ten classes (i.e., A-J) to match the D$^2$NN physical system, with 48000 training examples and 8000 testing examples. In the rest of this section, the total number of training iterations of all experiments is set to 150. Within each training iteration, each dataset is trained with 1 training epoch.
The learning rate in the training process is set to be $0.01$ for all experiments cross all four datasets using \texttt{Adam}. 
The implementations are constructed using \texttt{PyTorch v1.5.1}. All experiments are conducted on a Nvidia 2080 Ti GPU.
}

\subsection{Evaluations of RubikONNs with RotAgg and RotSeq Algorithms}\label{sec:result:comparison}

\begin{figure}[!htb]
\vspace{-1mm}
    \centering
    \begin{subfigure}[b]{0.9\linewidth}
    \centering
        \includegraphics[width=1\linewidth]{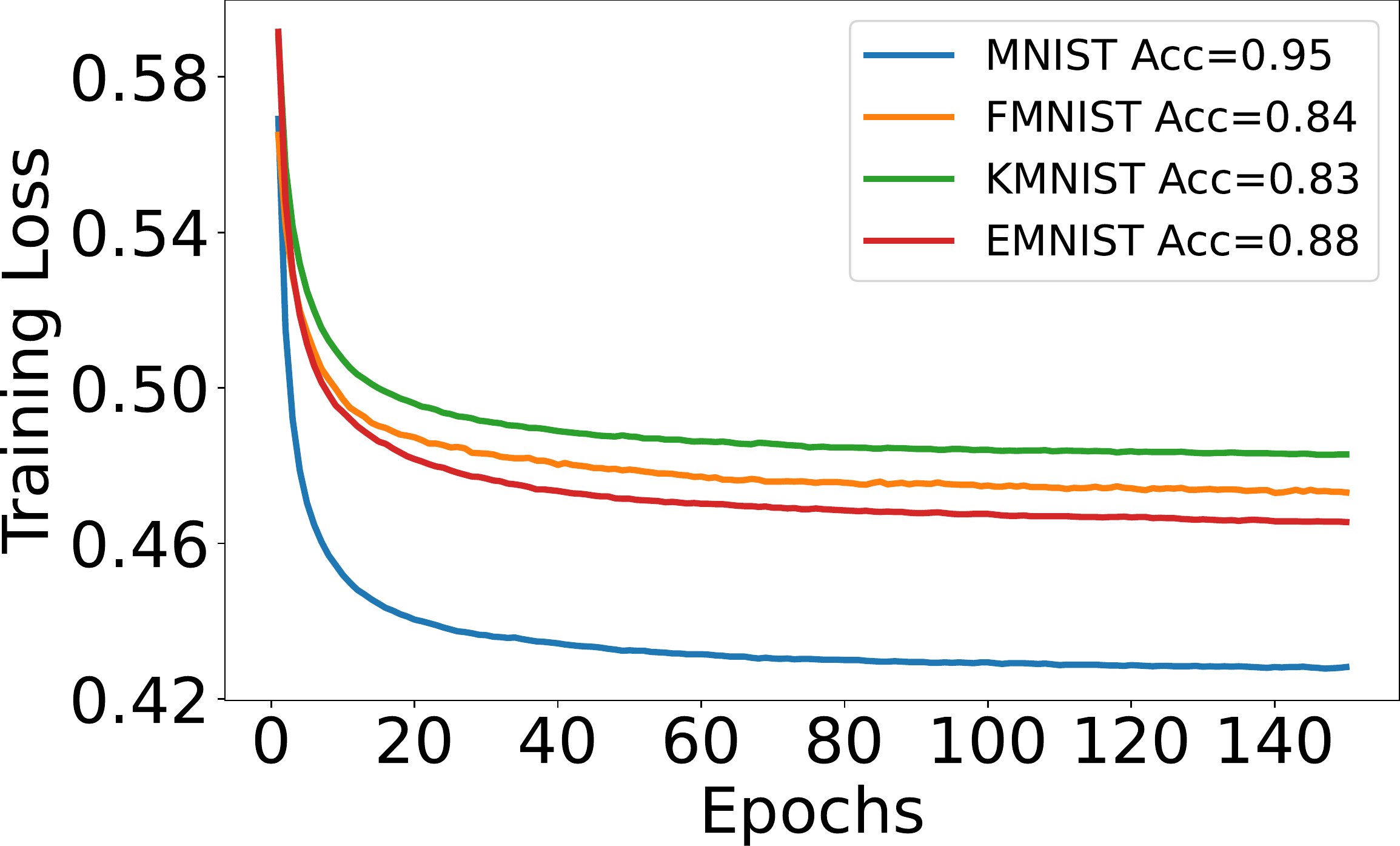}
        \caption{Training loss curve for Algorithm \ref{Alg: Averaging-weights}.}
        \label{fig:avg_loss}
    \end{subfigure}
    \hfill
    \begin{subfigure}[b]{0.9\linewidth}
    \centering
        \includegraphics[width=1\linewidth]{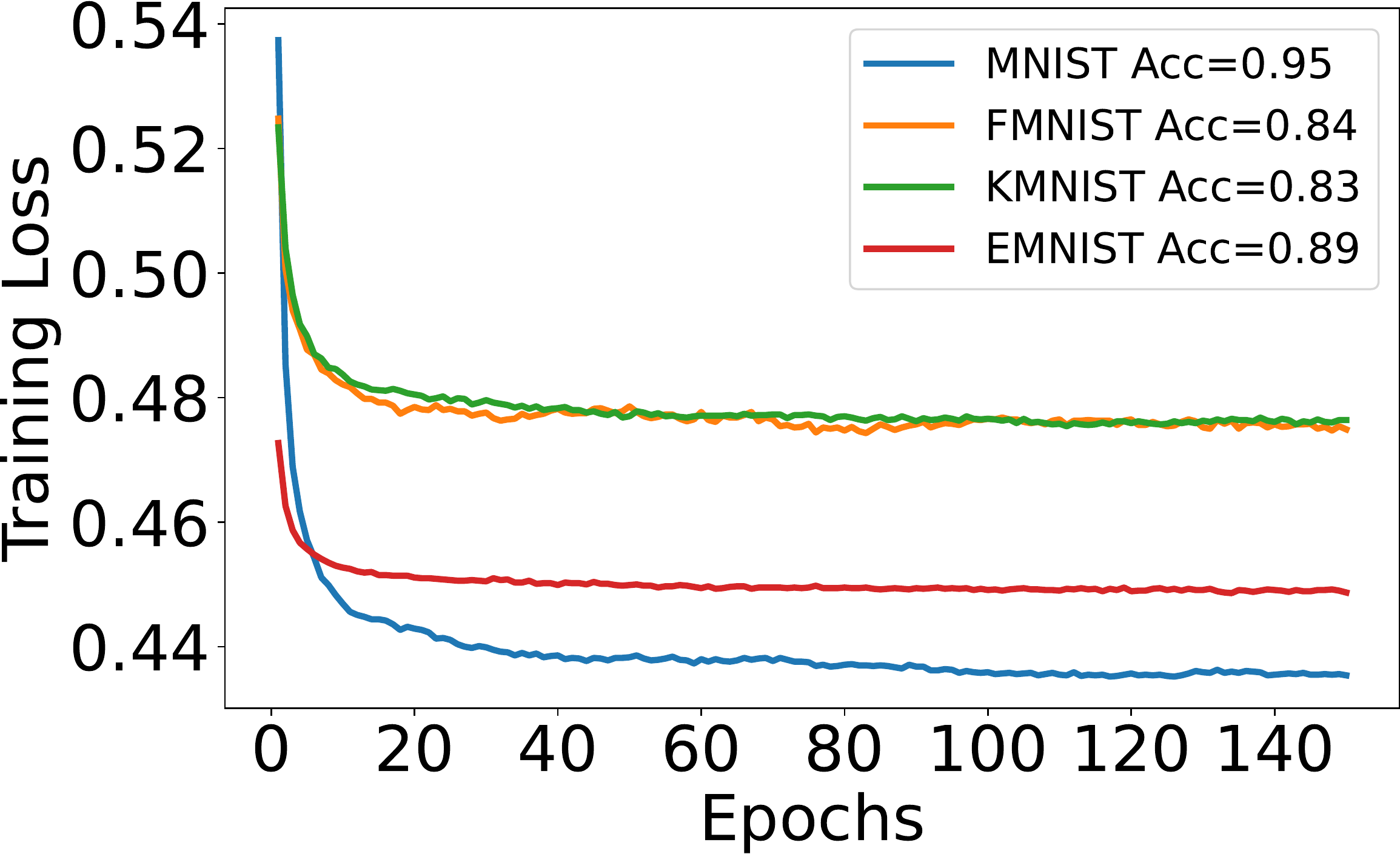}
        \caption{Training loss curve for Algorithm \ref{Alg: Sequencing-training} with the permutation D1$\shortrightarrow$D2$\shortrightarrow$D3$\shortrightarrow$D4.}
        \label{fig:seq_loss}
    \end{subfigure}
\caption{{Training loss curves with 150 epochs for two algorithms.}}
\vspace{-3mm}
\label{fig:eval_loss}
\end{figure}

\noindent\textbf{RotSeq and training permutations, and RotAgg} -- {We first evaluate RotSeq algorithm (Alg. \ref{Alg: Sequencing-training}) on MTL using the four selected datasets. As discussed in Section \ref{sec:approch}, the performance of RotSeq can vary with different gradient update orders (i.e., lines 4--10 in Alg. \ref{Alg: Sequencing-training}). Therefore, we evaluate RotSeq algorithm with four different permutations of the gradient update sequences, shown in second column of Table \ref{table:permutation_result}. With such recurring permutation of training orders, each task can be trained at each position in the training order. First, for each dataset, we can see that RotSeq offers a small accuracy boost for the given task/dataset used as the last gradient update in one RotSeq training iteration. This is because RotSeq (Alg. \ref{Alg: Sequencing-training}) updates the parameters in a given sequence to all training tasks, where testing accuracy is basically obtained right after the gradient updates of the last task.
Second, when a dataset is trained at the beginning of each training iteration (first task in the training sequence), the prediction performance of this task might slightly degrade. For example, MNIST accuracy collected using model trained with {D1 $\shortrightarrow$ D2 $\shortrightarrow$ D3 $\shortrightarrow$ {D4}} sequence is 0.0006/0.0006/0.0025 lower than the other three permutations.} 

{As for the comparison between RotAgg and RotSeq rotation training algorithms, the training loss curves for two algorithms are shown in Figure \ref{fig:eval_loss}. Both algorithms can converge efficiently and produce similarly decent accuracy performance, while the training loss for Algorithm \ref{Alg: Sequencing-training}, which trains the rotation MTL system with sequential datasets, shows more fluctuations.} As a result, the model trained with RotAgg algorithm shows overall performance and robustness since the training is not impacted by orders of gradient updates. Instead, RotAgg averages the gradients obtained independently for all tasks. The advantages of RotAgg can be summarized in two: \textbf{(1)} The training hyperparameter space is much limited than RotSeq since its performance is not influenced by the gradient update order; \textbf{(2)} The algorithm is expected to be more robust than RotSeq as RotSeq has slight training bias w.r.t the gradient update order. While RotSeq performs similarly to RotAgg with these four datasets, the bias training characteristics in RotSeq can potentially result in requiring more training setup exploration. Thus, in the rest of the result section, we use RotAgg as default algorithm for RubikONN architecture exploration analysis.

\noindent\textbf{Prediction performance comparisons} -- 
To fully demonstrate the effectiveness of the proposed approaches, we first compare the prediction performance with two existing approaches. First of all, a straightforward method to enable MTL on a fixed single-task D$^2$NNs architecture is to simply train a D$^2$NNs while merging the four datasets as one. Thus, we implement a straightforward baseline algorithm by extending the approach proposed by \cite{lin2018all}, where the training dataset consists of fully shuffled training samples from all four datasets, namely \textit{BaselineMTL}, and its depth is set to be five and system size is set to be $200\times200$, which is the same setup as our rotation system. The evaluation result of this baseline algorithm is shown in the last row of Table \ref{table:permutation_result}.
Next, we compare our approaches to a specific MTL D$^2$NNs architecture. Specifically, \cite{li2020multi} proposes a novel D$^2$NN architecture that utilizes transfer learning concepts from conventional neural networks, which includes shared diffractive layers (shared weights) and independent diffractive layers at the output stage for each task. To make a fair comparison, we extend that architecture to deal with four tasks and set three layers for the shared diffractive layers and two layers for the independent diffractive layers in each channel for four tasks and the same system size ($200\times200$) as our system.

As shown in Table \ref{table:permutation_result}, we can see that by utilizing the physical rotation properties with the proposed training algorithms, RubikONNs offers better prediction accuracy for all datasets. We can see that with RotAgg and RotSeq, RubikONNs performs significantly better than both baseline approaches. For example, with RotAgg algorithm, our approach offers about 2.5\% accuracy increases for MNIST and FMNIST, 6.5\% increases for EMNIST and 8.4\% for KMNIST, compared to \textit{BaselineMTL} (\cite{lin2018all}); compared to \cite{li2020multi}, RotAgg offers 3.5\% accuracy increases on EMNIST, and performs similarly for other three tasks. {This demonstrates that by utilizing the physical rotations into D$^2$NN architecture, RubikONNs offers clear prediction improvements over other approaches, while system cost, energy consumption, and complexity into the comparisons are not yet included in the comparisons.}

\begin{table*}
\centering
\captionsetup{justification=centering}
  \caption{Evaluations of hardware efficiency on multi-task learning compared with \protect\cite{lin2018all} and \protect\cite{li2020multi}, using datasets. MNIST(D1), FMNIST(D2), KMNIST(D3), and EMNIST(D4).}
    \label{table:Acc-HW Prod}
    \resizebox{1.7\columnwidth}{!}{%
\begin{tabular}{|c|c|c|c|c|c|c|c|c|c|c|c|c}
\hline
\multirow{2}{*}{} & \multicolumn{4}{c|}{\cite{lin2018all}} & \multicolumn{4}{c|}{\cite{li2020multi}} & \multicolumn{4}{c|}{\bf This work} \\ \cline{2-13} 
 & D1 & D2 & D3 & D4 & D1 & D2 & D3 & D4 & D1 & D2 & D3 & \multicolumn{1}{c|}{D4} \\ \hline
Layer Cost & 5 & 5 & 5 & 5 & \multicolumn{4}{c|}{11} & \multicolumn{4}{c|}{5} \\ \hline
Detector Cost* & 10 & 10 & 10 & 10 & \multicolumn{4}{c|}{20} & \multicolumn{4}{c|}{10} \\ \hline
\textit{Acc.(\%)} & 96.4 & 86.5 & 86.1 & 90.9 & 95.6 & 83.1 & 81.4 & 84.3 & 95.2 & 84.1 & 82.7 & \multicolumn{1}{c|}{88.2} \\ \hline
Cost Efficiency & 1 & 1 & 1 & 1 & 2.0{\scriptsize $\times$} & 2.1{\scriptsize $\times$} & 1.9{\scriptsize $\times$} & 1.9$  \times$ & {\bf 4.0{\scriptsize \ $\times$}} & {\bf 4.1{\scriptsize $\times$}} & {\bf 4.1{\scriptsize $\times$}} & \multicolumn{1}{c|}{{\bf 4.1{\scriptsize $\times$}}} \\ \hline
Power ($\mu W$/fps/task) & \multicolumn{4}{c|}{4.67$\times 10^{-7}$} & \multicolumn{4}{c|}{8.86$\times 10^{-7}$} & \multicolumn{4}{c|}{{\bf 1.7$\times 10^{-7}$}} \\ \hline
\end{tabular}
}
\vspace{-3mm}
\end{table*}

\noindent \textbf{Accuracy-Efficiency Comparison} -- To fully evaluate the efficiency of the models regarding the system cost, complexity, and energy efficiency, we introduce an accuracy-cost evaluation metric, where hardware cost is the sum of diffractive layer cost and detector cost\footnote{{The layer fabrication cost is $\sim$\$100 and detector cost is \$1,500 -- \$10,000. We formulate the cost of \$100 as unit 1, thus, the layer cost
for a 5-layer ONN is 5 and one detector cost is 10 for the cost efficiency estimation.}}. In Table \ref{table:Acc-HW Prod}, single-task cost metric is set as the baseline (unit 1), and the improvement of the architectures is calculated using Equation \ref{eq:acc-hw}. Note that in Table \ref{table:Acc-HW Prod}, the baseline results are collected using single-task implementation with five layers and $200\times200$ system size, and our results are generated using RotAgg algorithm. 
We can see that our approach offers \textbf{more than 4.0$\times$ and 2.0$\times$ hardware cost efficiency improvements} compared to \cite{lin2018all} and \cite{li2020multi}, respectively. Regarding energy efficiency, we evaluate the power consumption per task. Our approach demonstrates \textbf{2.7$\times$ and 5.3$\times$ energy efficiency improvements} compared to \cite{lin2018all} and \cite{li2020multi}, respectively. Note that the power consumption of DONNs is orders of magnitude more efficient than conventional digital platforms. Thus, we only compared to DONNs baselines in this work since the advantages of DONNs over conventional DNN hardware have been demonstrated.

\begin{equation} 
\begin{split}
\text{Acc-Efficiency Metric} &=  \frac{Acc_{\texttt{MTL}}}{Acc_{\texttt{baseline}}} \cdot \frac{Cost_{\texttt{MTL}}}{Cost_{\texttt{baseline}}};
\\
Cost &= \text{\#. Detectors or $\mu W$/fps/task}
\label{eq:acc-hw}
\vspace{-3mm}
\end{split}
\vspace{-5mm}
\end{equation}

\subsection{Design Space of RubikONNs Architecture}
With the proposed architecture and training algorithms, the rotation architecture can be designed in many different variants. Specifically, the rotation angles of the rotation layers, and the index of rotation layers to be rotated, which are independent to all other system and algorithm specifications. For example, instead of rotating the layers clockwise $90^\circ$, the layers can also be rotated $180^\circ$ and $270^\circ$ ($-90^\circ$). Similarly, the architecture can also be designed by selecting other layers other than the 4$^{th}$ and 5$^{th}$ layers to be rotated. Thus, in this section, we provide experimental analysis of other variants of the proposed architecture by evaluating different rotation angles and various rotation layer selections. {To limit the exploration space on the algorithm side for RubikONN architecture exploration, we explore the rotation angles and rotation layers only using \textbf{RotAgg} algorithm.}

\begin{table}
\small
\captionsetup{justification=centering}
  \caption{Explorations with various rotation angles (clockwise) with the 4$^{th}$ and 5$^{th}$ layers as rotation layers.}

  \centering
  \label{table:Rotation-angle-table}

  \begin{tabular}{ccccc}
  \toprule
    \multirow{2}{*}{\begin{tabular}[c]{@{}c@{}}Rotation \\ Angle\end{tabular}}  & \multicolumn{4}{c}{\textbf{RotAgg} (Alg. \ref{Alg: Averaging-weights}) w rot layers = 4$^{th}$, 5$^{th}$}   \\
    \cmidrule(r){2-5}
      &  MNIST & FMNIST & KMNIST & EMNIST\\
    \midrule
    $90^\circ$, $90^\circ$   & {0.9524} & {0.8412} & {0.8272} & {0.8819}    \\
    $180^\circ$, $180^\circ$  &{0.9532} &{0.8514} & {0.8313} & {0.8809}   \\
    $270^\circ$, $270^\circ$  & {0.9527} & {0.8353} & {0.8240} & {0.8845}    \\
    $90^\circ$, $-90^\circ$  & {0.9518} & {0.8427} & {0.8272} & {0.8879}    \\
    $90^\circ$, $180^\circ$  & {0.9514} & {0.8413} & {0.8227} & {0.8851}    \\
    \bottomrule
  \end{tabular}
\end{table}

\begin{table}

\small
\captionsetup{justification=centering}
  \caption{Design space explorations with different selections of rotation layers with rotation angle $90^\circ$.}
  \centering
  \label{table:Rotation-layer-table}
  \begin{tabular}{ccccc}
    \toprule
    \multirow{2}{*}{\begin{tabular}[c]{@{}c@{}}Rotated\\Layers\end{tabular}}  &   \multicolumn{4}{c}{\textbf{RotAgg} (Alg. \ref{Alg: Averaging-weights})}   \\
    \cmidrule(r){2-5}
     &  MNIST & FMNIST &KMNIST &EMNIST\\
    \midrule
    4$^{th}$, 5$^{th}$      &{0.9524} & {0.8412} & {0.8272} & {0.8819}    \\
    1$^{st}$, 2$^{nd}$      &{0.9536} &{0.8490} & {0.8199} & {0.8813}  \\
    1$^{st}$, 5$^{th}$       &{0.9510} & {0.8473} & {0.8210} & {0.8865}\\
    \bottomrule
  \end{tabular}

\end{table}

\noindent \textbf{Analysis of different rotation angles} -- {Since each diffractive layer can rotate close-wise $90^\circ$, $180^\circ$, and $270^\circ$ ($-90^\circ$), the rotation angle can be independent from layer to layer, e.g., rotating 4$^{th}$ layer $90^\circ$ and rotating 5$^{th}$ layer by $180^\circ$. To evaluate the impacts of rotation angles, the experiments shown in Table \ref{table:Rotation-angle-table} are conducted with fixed selection of rotation layers, i.e., 4$^{th}$ and 5$^{th}$ layers. Table \ref{table:Rotation-angle-table} shows the accuracy of four datasets in the model trained with RotAgg when different rotation angles are applied to last two layers. Specifically, we evaluate two different rotation angle settings: (1) same rotation angles for both layers; or (2) different rotation angles for the two layers. For example, ($90^\circ$, $180^\circ$) means that if the 4$^{th}$ layer is designed to be rotated for a given task, it rotates $90^\circ$; and 5$^{th}$ layer is rotated $180^\circ$ if needed. In general, with different rotation angles, RubikONNs shows little fluctuation in terms of accuracy.} 

\noindent \textbf{Analysis of rotated layer selections} -- {
Let the number of tasks be 4 and each rotation layer can only rotate clock-wise $90^\circ$, the total number of layer selections is $\textbf{C}^{2}_{5}$=$10$. According to studies of conventional neural networks, the layers close to the inputs are usually very important for feature extractions, while the layers close to outputs are crucial for generating the final prediction class. Thus, we evaluate three combinations, including (1) the last two layers, (2) the first two layers, and (3) the first and the last layers. The results are shown in Table \ref{table:Rotation-layer-table}. We can see that (a) the models trained with RotAgg algorithm perform almost the same, regardless of which layers are selected as rotation layers; (2) including the last layer (5$^{th}$) in the rotation layers performs slightly better on average. 
} 

In summary, Table \ref{table:Rotation-angle-table} and Table \ref{table:Rotation-layer-table} results suggest the follows: \textbf{(1)} The rotation layers are preferred to be selected close to the output. \textbf{(2)} The prediction performance is not restricted to specific rotation angles, which offers possibility to encode more forward functions, and it is the key to enable larger number of tasks for MTL. 

\subsection{RubikONNs MTL Explainability}

\begin{figure}[!htb]
    \centering
\subcaptionbox{Evaluating MNIST "1" with model designed for MNIST (rotation pattern [$90^\circ$, $0^\circ$]). }
{\includegraphics[width=1\linewidth]{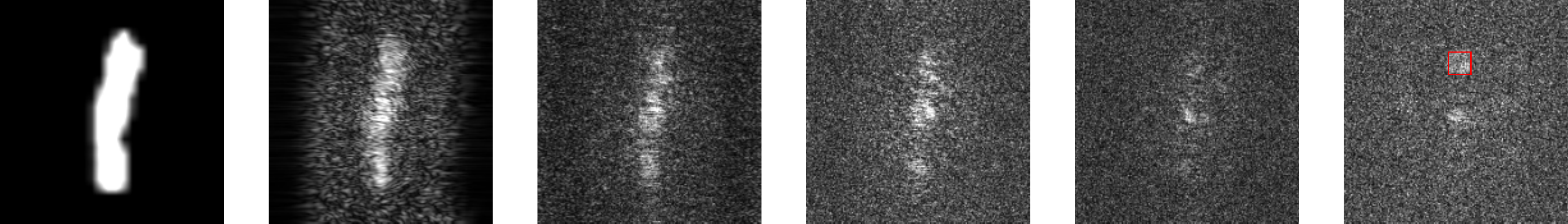}} \\
\subcaptionbox{Evaluating MNIST "1" with model designed for FMNIST (rotation pattern [$0^\circ$, $0^\circ$]). }
{\includegraphics[width=1\linewidth]{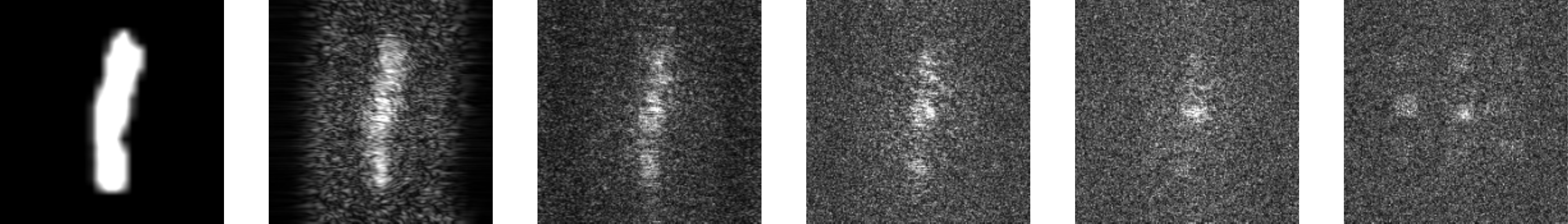}} \\
\subcaptionbox{Evaluating MNIST "1" with model designed for KMNIST (rotation pattern [$90^\circ$, $90^\circ$]). }
{\includegraphics[width=1\linewidth]{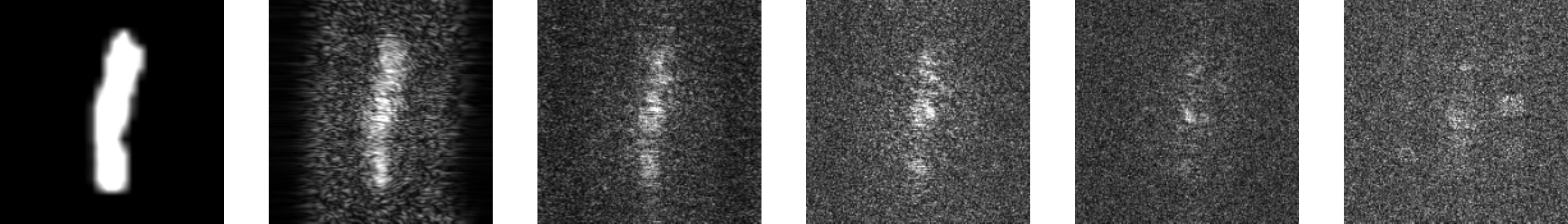}} \\
\subcaptionbox{Evaluating MNIST "1" with model designed for EMNIST (rotation pattern [$0^\circ$, $90^\circ$]). }
{\includegraphics[width=1\linewidth]{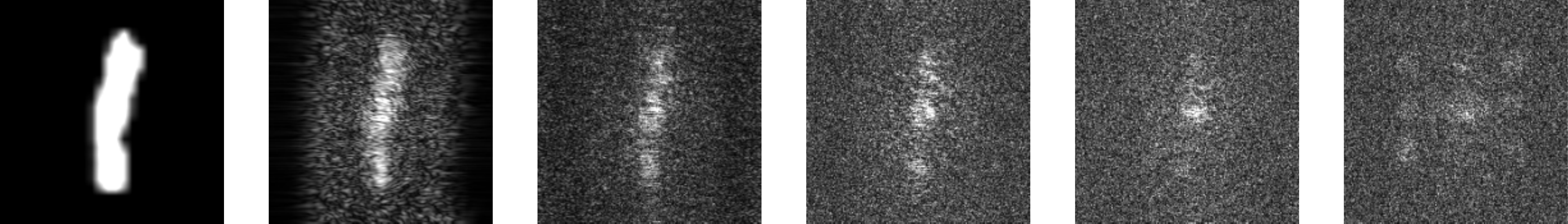}} \\
    \caption{Visualization of light propagation patterns at inference with MNIST image "1" as input, using all four rotations of RubikONNs.
    }
    \vspace{-3mm}
    \label{fig:prop-mnist}
\end{figure}

\begin{figure}
    \centering
\includegraphics[width=0.31\linewidth]{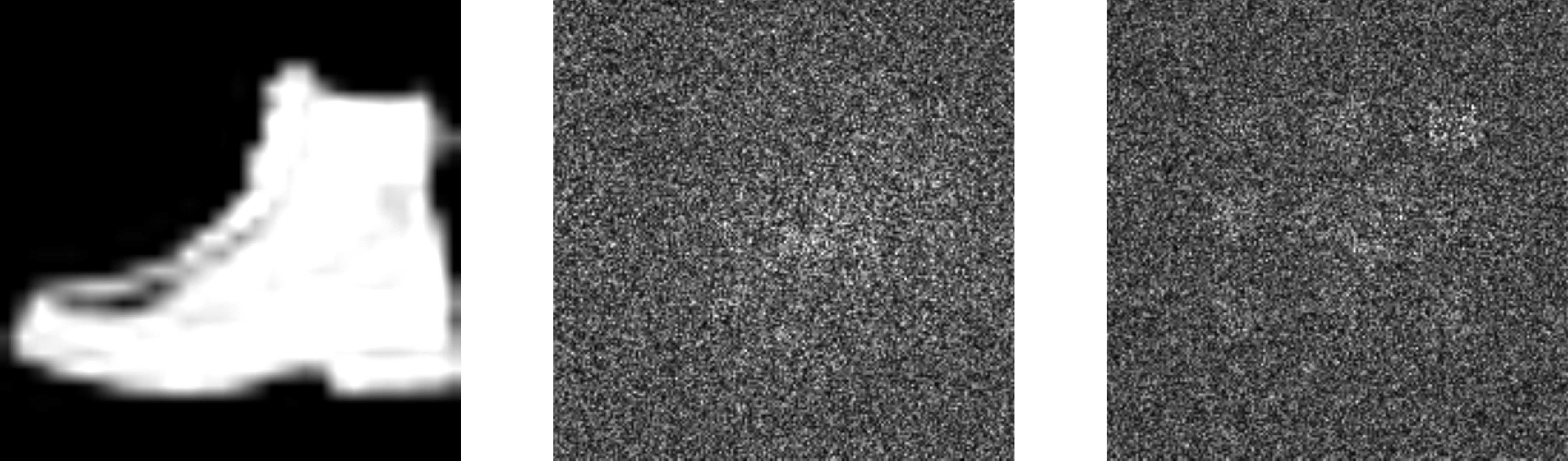}
\includegraphics[width=0.31\linewidth]{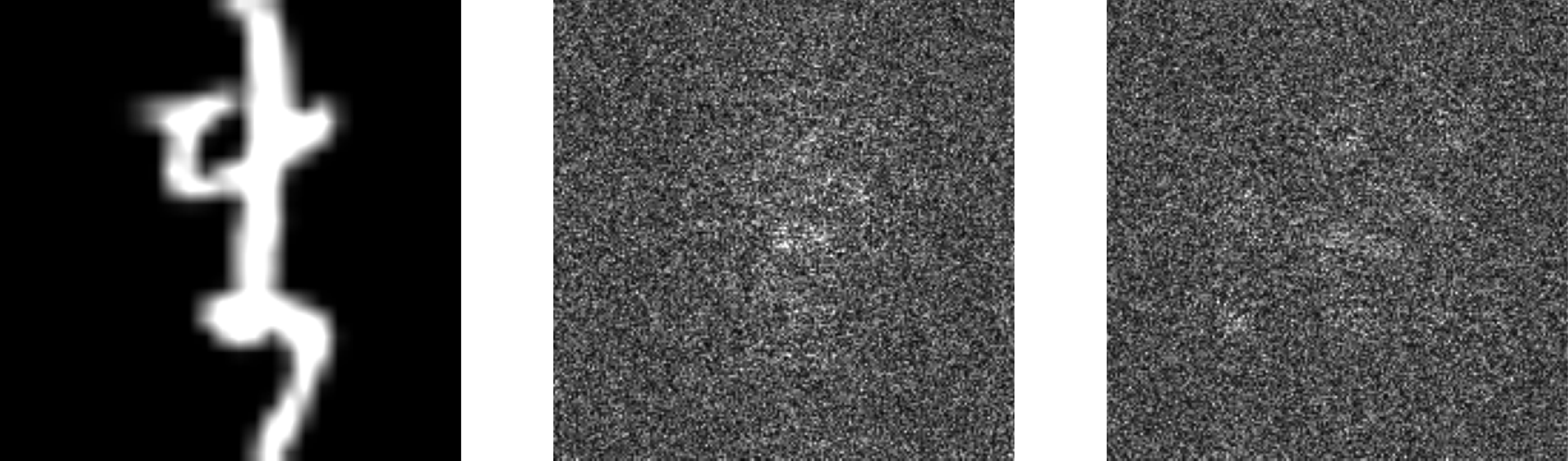} 
\includegraphics[width=0.31\linewidth]{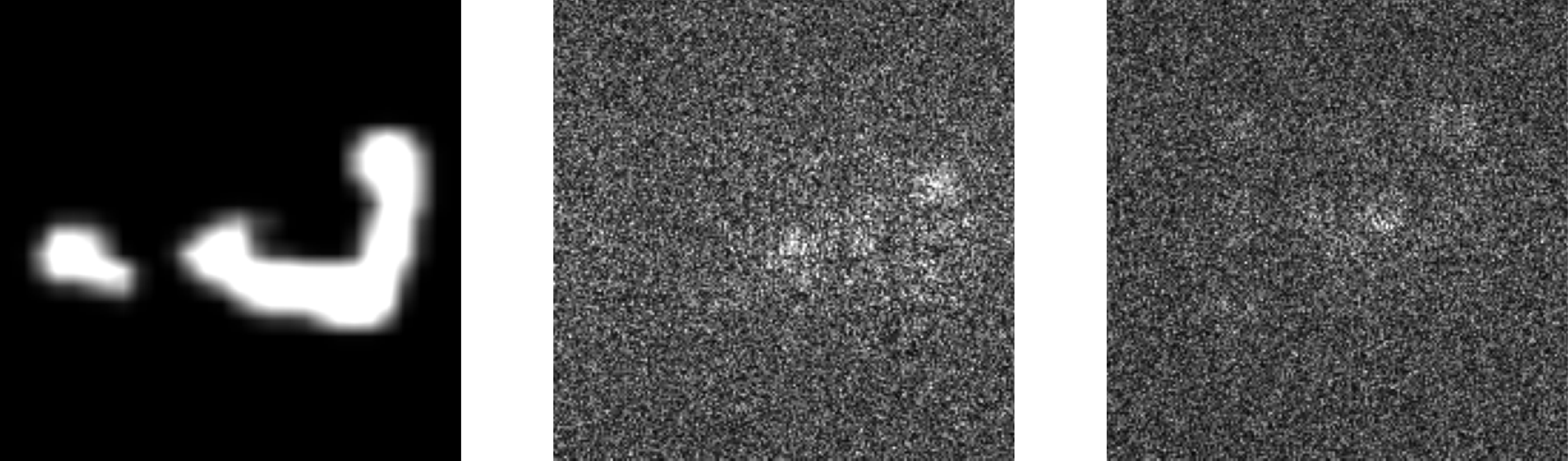}
\\
\includegraphics[width=0.31\linewidth]{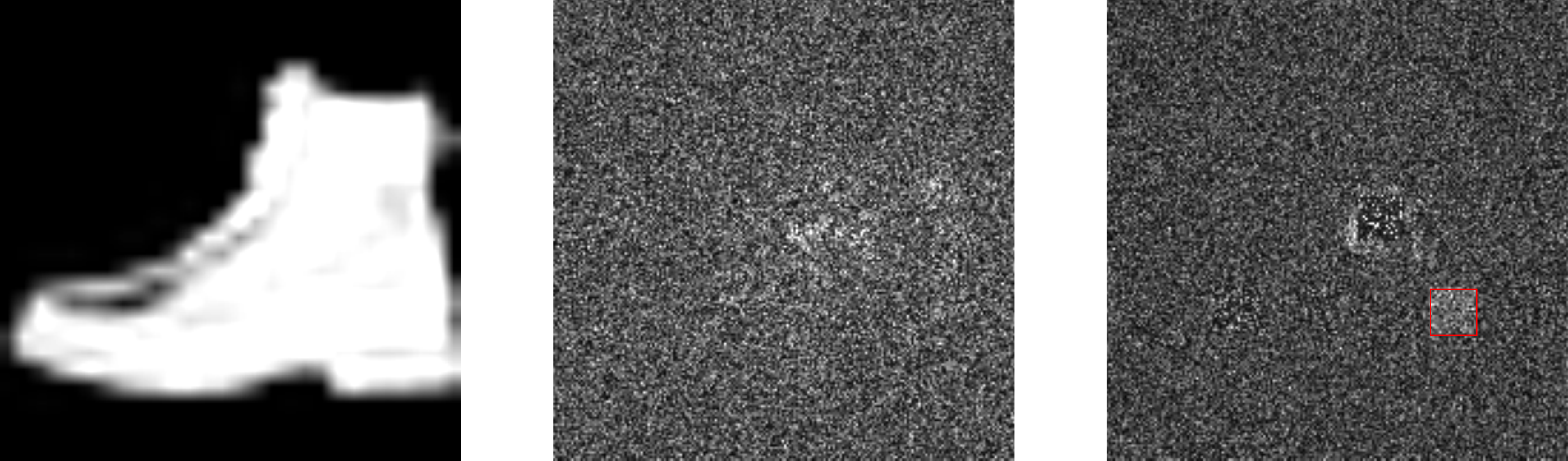}
\includegraphics[width=0.31\linewidth]{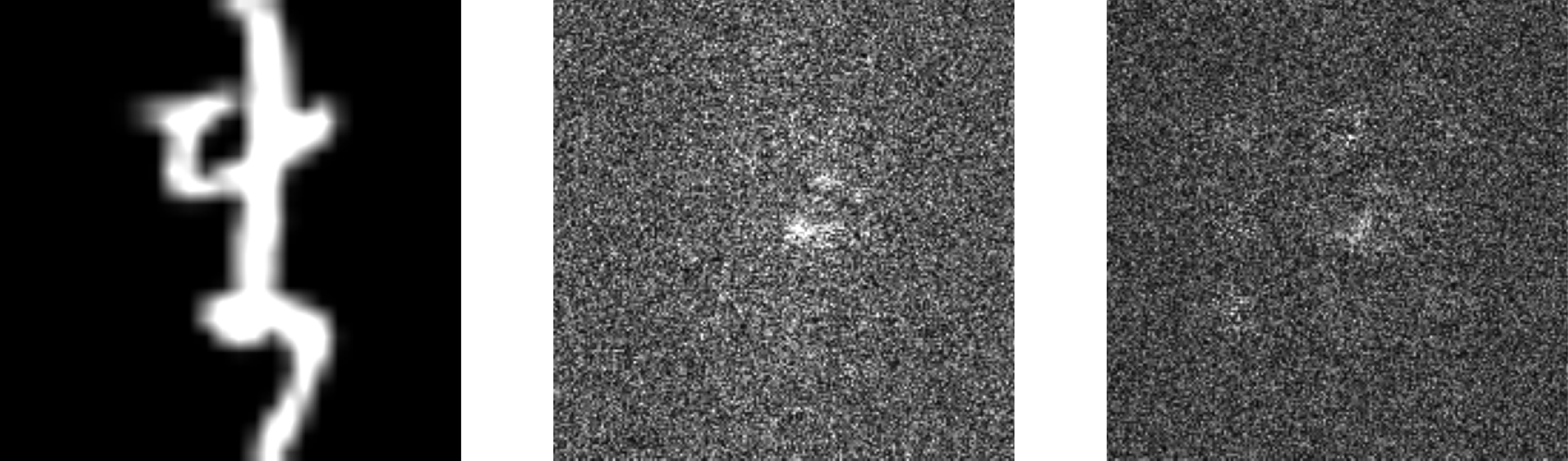} 
\includegraphics[width=0.31\linewidth]{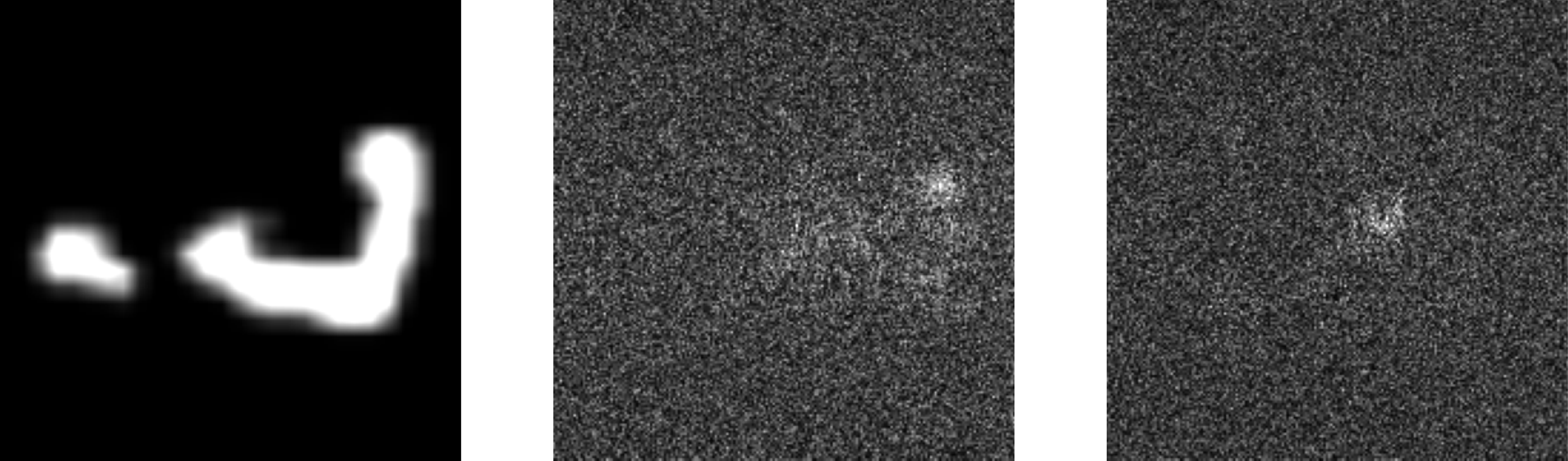}
\\
\includegraphics[width=0.31\linewidth]{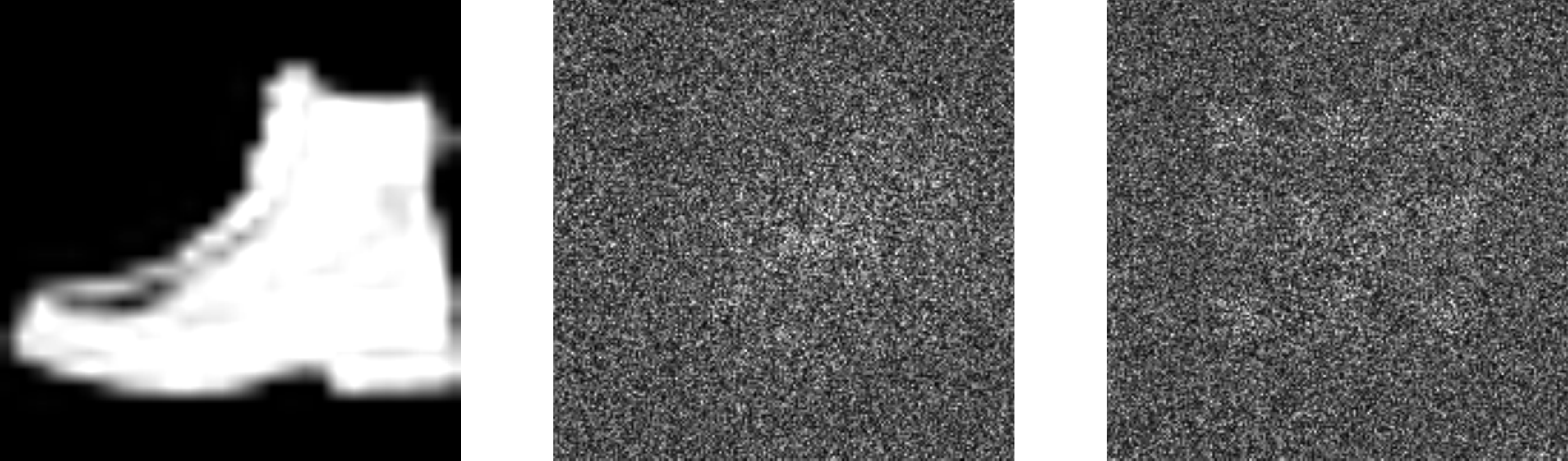}
\includegraphics[width=0.31\linewidth]{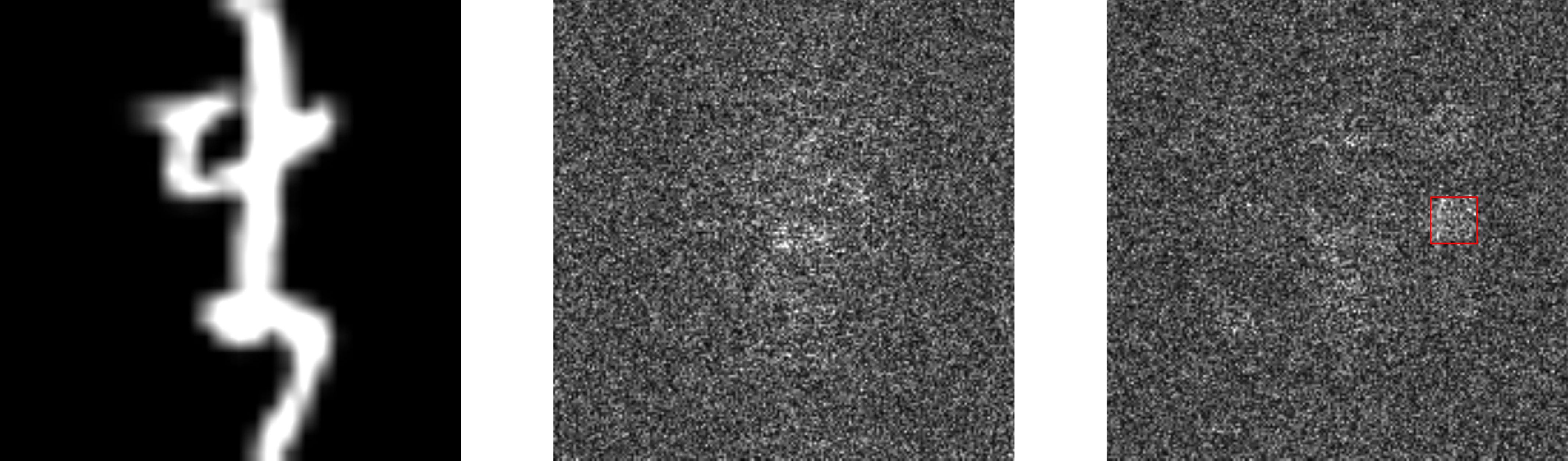} 
\includegraphics[width=0.31\linewidth]{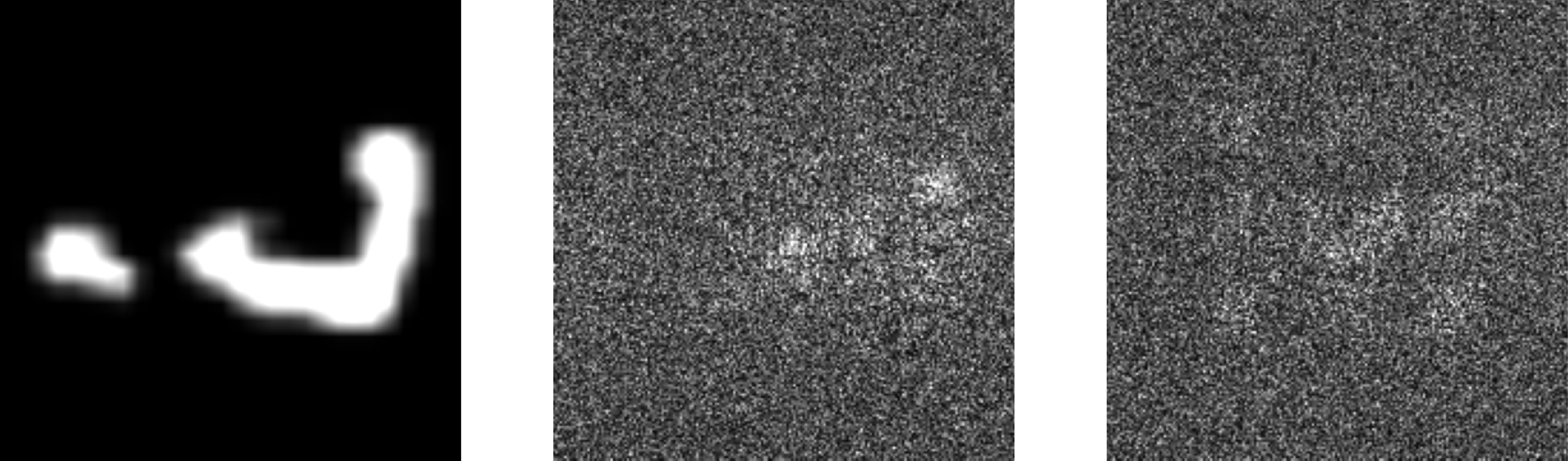}
\\
\subcaptionbox{FMNIST label ''ankle boot''.}
{\includegraphics[width=0.31\linewidth]{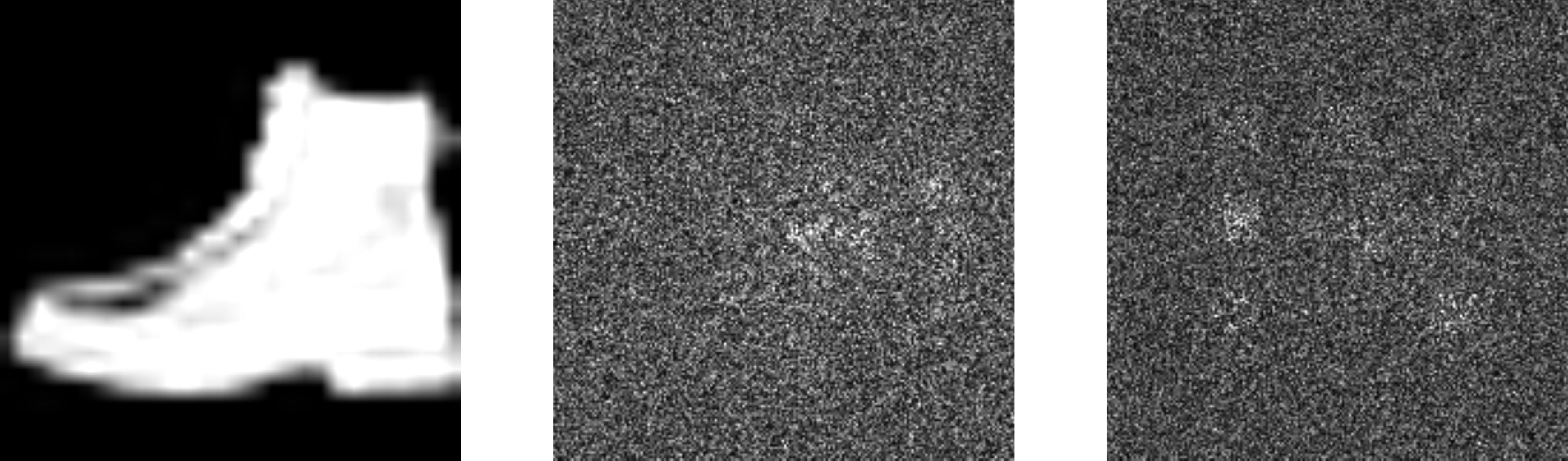}}
\subcaptionbox{KMNIST with label \begin{CJK}{UTF8}{min}
''き''
\end{CJK}. }
{\includegraphics[width=0.31\linewidth]{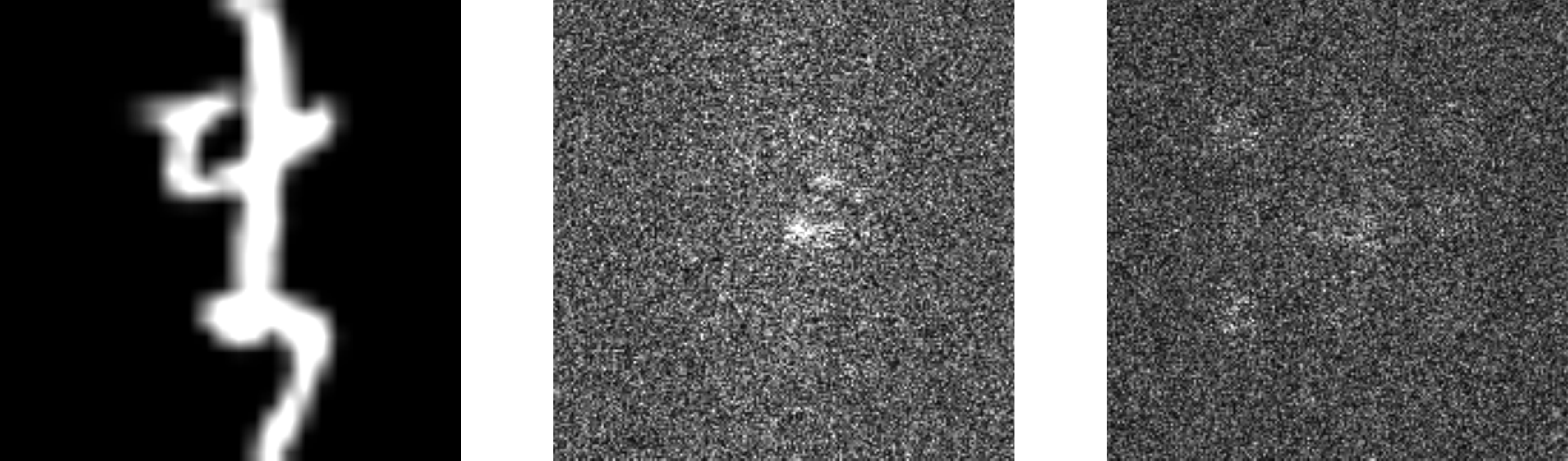}} 
\subcaptionbox{EMNIST with label ''j''.}
{\includegraphics[width=0.31\linewidth]{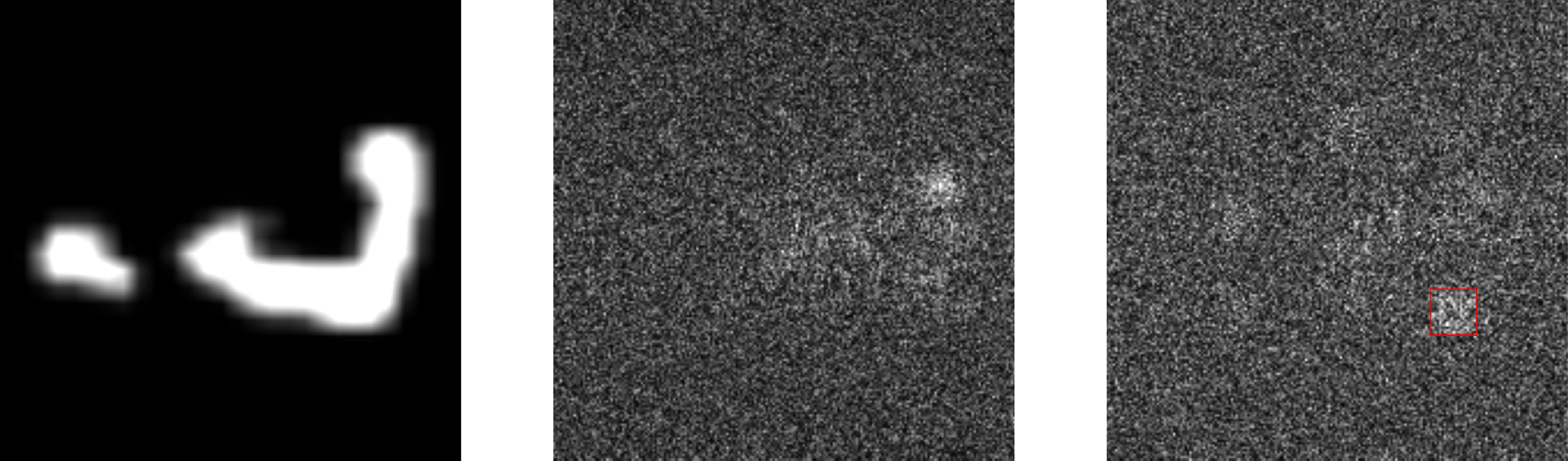}}
\\    
    \caption{Visualization of light propagation measurements (input, and outputs of 4$^{th}$ and 5$^{th}$ layers) at inference phase with FMNIST, KMNIST, and EMNIST images, using all four rotations of RubikONNs. }
    \vspace{-3mm}
    \label{fig:prop-others}
\end{figure}

To understand the impacts of rotations for MTL, we measure the internal propagation of RubikONNs between the source, layers and detectors. Specifically, we measure the intensity of the light in the RubikONNs at inference phase, shown in Figures \ref{fig:prop-mnist}. The visualizations of the forward propagation shown in Figures \ref{fig:prop-mnist} are organized by applying a same image from one dataset using all rotation patterns, following the designed rotation patterns shown in Figure \ref{fig:rotation}. It is known that the main idea of DNNs is that layers close to the input focus on extracting features, and layers close to the output focus on finalizing the predictions using the extracted features. The intuition of RubikONNs architecture is relatively the same, and has been demonstrated based on the propagation measurements. For example, in Figure \ref{fig:prop-mnist}, the input image is from MNIST dataset, where four complete propagation measurements are included w.r.t the rotation patterns for task MNIST, FMNIST, KMNIST, and EMNIST, respectively. We can see that the outputs of the first three layers are identical for all four cases, since the first three layers are not the rotation layers. The differences of forward propagation are observed starting from the 4$^{th}$ layer, which is rotated clockwise $90^\circ$ for MNIST and KMNIST tasks, and remains un-rotated for FMNIST and EMNIST tasks. Similarly, since the 5$^{th}$ layer is also designed to be rotated as well, the outputs collected by the detectors clearly show four different intensity distributions. { Additional outputs of 4$^{th}$ and 5$^{th}$ layers w.r.t other three datasets are shown in Figure \ref{fig:prop-others}, which further confirms that} RubikONNs is able to successfully encode four different forward functions, which are properly optimized for four tasks using the proposed training algorithms.  

\section{Conclusions}

This work proposes a novel optical neural architecture \textbf{RubikONNs} architecture, which utilizes the physical properties of optical computing systems to encode multiple feed-forward functions by rotating the non-reconfigurable hardware system. To optimize the MTL performance of RubikONNs, two novel domain-specific physics-aware training algorithms {RotAgg} and {RotSeq} are proposed, such that RubikONNs offers \textbf{4$\mathbf{\times}$} implementation cost and energy efficiencies improvements, with marginal accuracy degradation. Finally, a comprehensive RubikONNs design space exploration analysis and explainability are provided to offer concrete design methodologies for practical uses. 

{The ONNs have the potential to handle more complex image tasks, including image classification tasks and graph tasks \cite{yan2022all} with outstanding energy efficiency advantage over conventional NNs (e.g., CNNs), which can already be observed with simple datasets like MNIST ($\sim$3 orders compared to CPU/GPU in our setups). It is believed that more significant improvements can be achieved when dealing with more complex tasks requiring more computation resources, as there is no extra energy cost for computations with the modulations of the light signal in ONNs.}

Future works will focus on dealing with more complex datasets. For example, we can do MTL with RubikONNs for RGB image classification by extending the optical channels for "R", "G", "B" channels with rotation scheme in each channel to realize the MTL for RGB images. 

\smallskip
\noindent \textbf{Acknowledgement} This work is funded by National Science Foundation (NSF) under NSF-2008144, NSF-2019306, NSF-2019336, and NSF-2047176.

%
\newpage
\bibliographystyle{named}
\bibliography{bibs/dlsys_yu.bib,bibs/analog.bib,bibs/adv.bib}

\begin{thebibliography}{}

\bibitem[\protect\citeauthoryear{Chen \bgroup \em et al.\egroup
  }{2022a}]{chen2022physics}
Ruiyang Chen, Yingjie Li, Minhan Lou, Jichao Fan, Yingheng Tang, Berardi
  Sensale-Rodriguez, Cunxi Yu, and Weilu Gao.
\newblock Physics-aware machine learning and adversarial attack in
  complex-valued reconfigurable diffractive all-optical neural network.
\newblock {\em Laser \& Photonics Reviews}, page 2200348, 2022.

\bibitem[\protect\citeauthoryear{Chen \bgroup \em et al.\egroup
  }{2022b}]{chen2022complex}
Ruiyang Chen, Yingjie Li, Minhan Lou, Cunxi Yu, and Weilu Gao.
\newblock Complex-valued reconfigurable diffractive optical neural networks
  using cost-effective spatial light modulators.
\newblock In {\em CLEO: Applications and Technology}, pages JTh3B--56. Optica
  Publishing Group, 2022.

\bibitem[\protect\citeauthoryear{Clanuwat \bgroup \em et al.\egroup
  }{2018}]{clanuwat2018deep}
Tarin Clanuwat, Mikel Bober-Irizar, Asanobu Kitamoto, Alex Lamb, Kazuaki
  Yamamoto, and David Ha.
\newblock Deep learning for classical japanese literature.
\newblock 2018.

\bibitem[\protect\citeauthoryear{Cohen \bgroup \em et al.\egroup
  }{2017}]{cohen2017emnist}
Gregory Cohen, Saeed Afshar, Jonathan Tapson, and Andre Van~Schaik.
\newblock Emnist: Extending mnist to handwritten letters.
\newblock In {\em 2017 International Joint Conference on Neural Networks
  (IJCNN)}, pages 2921--2926. IEEE, 2017.

\bibitem[\protect\citeauthoryear{Duan \bgroup \em et al.\egroup
  }{2023}]{duan2023optical}
Zhengyang Duan, Hang Chen, and Xing Lin.
\newblock Optical multi-task learning using multi-wavelength diffractive deep
  neural networks.
\newblock {\em Nanophotonics}, 2023.

\bibitem[\protect\citeauthoryear{Feldmann \bgroup \em et al.\egroup
  }{2019}]{feldmann2019all}
J~Feldmann, N~Youngblood, C~David Wright, H~Bhaskaran, and WHP Pernice.
\newblock All-optical spiking neurosynaptic networks with self-learning
  capabilities.
\newblock {\em Nature}, 569(7755):208--214, 2019.

\bibitem[\protect\citeauthoryear{Gao \bgroup \em et al.\egroup
  }{2021}]{gao2021artificial}
Weilu Gao, Cunxi Yu, and Ruiyang Chen.
\newblock Artificial intelligence accelerators based on graphene optoelectronic
  devices.
\newblock {\em Advanced Photonics Research}, 2(6):2100048, 2021.

\bibitem[\protect\citeauthoryear{Gu \bgroup \em et al.\egroup
  }{2020}]{gu2020towards}
Jiaqi Gu, Zheng Zhao, Chenghao Feng, Mingjie Liu, Ray~T Chen, and David~Z Pan.
\newblock Towards area-efficient optical neural networks: an fft-based
  architecture.
\newblock In {\em 2020 25th Asia and South Pacific Design Automation Conference
  (ASP-DAC)}, pages 476--481. IEEE, 2020.

\bibitem[\protect\citeauthoryear{Gu \bgroup \em et al.\egroup
  }{2022}]{gu2022adept}
Jiaqi Gu, Hanqing Zhu, Chenghao Feng, Zixuan Jiang, Mingjie Liu, Shuhan Zhang,
  Ray~T Chen, and David~Z Pan.
\newblock Adept: Automatic differentiable design of photonic tensor cores.
\newblock In {\em Proceedings of the 59th ACM/IEEE Design Automation
  Conference}, pages 937--942, 2022.

\bibitem[\protect\citeauthoryear{Jouppi \bgroup \em et al.\egroup
  }{2017}]{jouppi2017datacenter}
Norman~P Jouppi, Cliff Young, Nishant Patil, David Patterson, Gaurav Agrawal,
  Raminder Bajwa, Sarah Bates, Suresh Bhatia, Nan Boden, Al~Borchers, et~al.
\newblock {In-datacenter Performance Analysis of a Tensor Processing Unit}.
\newblock {\em Int'l Symp. on Computer Architecture (ISCA)}, pages 1--12, 2017.

\bibitem[\protect\citeauthoryear{LeCun}{1998}]{lecun1998mnist}
Yann LeCun.
\newblock The mnist database of handwritten digits.
\newblock {\em http://yann. lecun. com/exdb/mnist/}, 1998.

\bibitem[\protect\citeauthoryear{Li and Yu}{2021}]{li2021late}
Yingjie Li and Cunxi Yu.
\newblock Late breaking results: physical adversarial attacks of diffractive
  deep neural networks.
\newblock In {\em DAC}, 2021.

\bibitem[\protect\citeauthoryear{Li \bgroup \em et al.\egroup
  }{2021}]{li2020multi}
Yingjie Li, Ruiyang Chen, Berardi~Sensale Rodriguez, Weilu Gao, and Cunxi Yu.
\newblock Multi-task learning in diffractive deep neural networks via
  hardware-software co-design.
\newblock {\em Scientific Reports}, pages 1--9, 2021.

\bibitem[\protect\citeauthoryear{Li \bgroup \em et al.\egroup
  }{2022}]{li2022physics}
Yingjie Li, Ruiyang Chen, Weilu Gao, and Cunxi Yu.
\newblock Physics-aware differentiable discrete codesign for diffractive
  optical neural networks.
\newblock In {\em Proceedings of the 41st IEEE/ACM International Conference on
  Computer-Aided Design}, pages 1--9, 2022.

\bibitem[\protect\citeauthoryear{Lin \bgroup \em et al.\egroup
  }{2018}]{lin2018all}
Xing Lin, Yair Rivenson, Nezih~T Yardimci, Muhammed Veli, Yi~Luo, Mona Jarrahi,
  and Aydogan Ozcan.
\newblock All-optical machine learning using diffractive deep neural networks.
\newblock {\em Science}, 361(6406):1004--1008, 2018.

\bibitem[\protect\citeauthoryear{Lou and \textit{et
  al.}}{2023}]{lou2023effects}
Minhan Lou and \textit{et al.}
\newblock Effects of interlayer reflection and interpixel interaction in
  diffractive optical neural networks.
\newblock {\em Optics Letters}, 2023.

\bibitem[\protect\citeauthoryear{Mengu \bgroup \em et al.\egroup
  }{2020}]{mengu2020scale}
Deniz Mengu, Yair Rivenson, and Aydogan Ozcan.
\newblock Scale-, shift-and rotation-invariant diffractive optical networks.
\newblock {\em arXiv preprint arXiv:2010.12747}, 2020.

\bibitem[\protect\citeauthoryear{Mengu \bgroup \em et al.\egroup
  }{2023}]{mengu2023snapshot}
Deniz Mengu, Anika Tabassum, Mona Jarrahi, and Aydogan Ozcan.
\newblock Snapshot multispectral imaging using a diffractive optical network.
\newblock {\em Light: Science \& Applications}, 12(1):86, 2023.

\bibitem[\protect\citeauthoryear{Rahman \bgroup \em et al.\egroup
  }{2020}]{rahman2020ensemble}
Md~Sadman~Sakib Rahman, Jingxi Li, Deniz Mengu, Yair Rivenson, and Aydogan
  Ozcan.
\newblock Ensemble learning of diffractive optical networks.
\newblock {\em arXiv preprint arXiv:2009.06869}, 2020.

\bibitem[\protect\citeauthoryear{Shen \bgroup \em et al.\egroup
  }{2017}]{shen2017deep}
Yichen Shen, Nicholas~C Harris, Scott Skirlo, Mihika Prabhu, Tom Baehr-Jones,
  Michael Hochberg, Xin Sun, Shijie Zhao, Hugo Larochelle, Dirk Englund, et~al.
\newblock Deep learning with coherent nanophotonic circuits.
\newblock {\em Nature Photonics}, 11(7):441, 2017.

\bibitem[\protect\citeauthoryear{Strubell \bgroup \em et al.\egroup
  }{2019}]{strubell2019energy}
Emma Strubell, Ananya Ganesh, and Andrew McCallum.
\newblock Energy and policy considerations for deep learning in nlp.
\newblock {\em arXiv preprint arXiv:1906.02243}, 2019.

\bibitem[\protect\citeauthoryear{Tait \bgroup \em et al.\egroup
  }{2017}]{tait2017neuromorphic}
Alexander~N Tait, Thomas~Ferreira De~Lima, Ellen Zhou, Allie~X Wu, Mitchell~A
  Nahmias, Bhavin~J Shastri, and Paul~R Prucnal.
\newblock Neuromorphic photonic networks using silicon photonic weight banks.
\newblock {\em Scientific reports}, 7(1):1--10, 2017.

\bibitem[\protect\citeauthoryear{Tang \bgroup \em et al.\egroup
  }{2023}]{tang2023device}
Yingheng Tang, Princess~Tara Zamani, Ruiyang Chen, Jianzhu Ma, Minghao Qi,
  Cunxi Yu, and Weilu Gao.
\newblock Device-system end-to-end design of photonic neuromorphic processor
  using reinforcement learning.
\newblock {\em Laser \& Photonics Reviews}, 17(2):2200381, 2023.

\bibitem[\protect\citeauthoryear{Xiao \bgroup \em et al.\egroup
  }{2017}]{xiao2017fashion}
Han Xiao, Kashif Rasul, and Roland Vollgraf.
\newblock Fashion-mnist: a novel image dataset for benchmarking machine
  learning algorithms.
\newblock {\em arXiv preprint arXiv:1708.07747}, 2017.

\bibitem[\protect\citeauthoryear{Yan \bgroup \em et al.\egroup
  }{2022}]{yan2022all}
Tao Yan, Rui Yang, Ziyang Zheng, Xing Lin, Hongkai Xiong, and Qionghai Dai.
\newblock All-optical graph representation learning using integrated
  diffractive photonic computing units.
\newblock {\em Science Advances}, 8(24):eabn7630, 2022.

\bibitem[\protect\citeauthoryear{Yazdanbakhsh \bgroup \em et al.\egroup
  }{2021}]{yazdanbakhsh2021evaluation}
Amir Yazdanbakhsh, Kiran Seshadri, Berkin Akin, James Laudon, and Ravi
  Narayanaswami.
\newblock An evaluation of edge tpu accelerators for convolutional neural
  networks.
\newblock {\em arXiv e-prints}, pages arXiv--2102, 2021.

\bibitem[\protect\citeauthoryear{Yin \bgroup \em et al.\egroup
  }{2022}]{yin2022exact}
Jiaqi Yin, Zhiru Zhang, and Cunxi Yu.
\newblock Exact memory-and communication-aware scheduling of dnns on pipelined
  edge tpus.
\newblock In {\em 2022 IEEE/ACM 7th Symposium on Edge Computing (SEC)}, pages
  203--215. IEEE, 2022.

\bibitem[\protect\citeauthoryear{Yin \bgroup \em et al.\egroup
  }{2023}]{yin2023respect}
Jiaqi Yin, Yingjie Li, Daniel Robinson, and Cunxi Yu.
\newblock Respect: Reinforcement learning based edge scheduling on pipelined
  coral edge tpus.
\newblock {\em Design Automation Conference (DAC'23)}, 2023.

\bibitem[\protect\citeauthoryear{Ying \bgroup \em et al.\egroup
  }{2020}]{ying2020electronic}
Zhoufeng Ying, Chenghao Feng, Zheng Zhao, Shounak Dhar, Hamed Dalir, Jiaqi Gu,
  Yue Cheng, Richard Soref, David~Z Pan, and Ray~T Chen.
\newblock Electronic-photonic arithmetic logic unit for high-speed computing.
\newblock {\em Nature communications}, 2020.

\end{thebibliography}

\end{document}